\documentclass[journal]{IEEEtran}
\usepackage{amsmath,amsfonts}
\usepackage{algorithmic}
\usepackage{algorithm}
\usepackage{array}
\usepackage[caption=false,font=normalsize,labelfont=sf,textfont=sf]{subfig}
\usepackage{textcomp}
\usepackage{stfloats}
\usepackage{url}
\usepackage{verbatim}
\usepackage{graphicx}
\usepackage{cite}
\usepackage{bm}
\usepackage{booktabs} 
\usepackage{multirow}
\usepackage{hyperref}
\usepackage{hhline}
\usepackage{xcolor} 
\usepackage{pifont}
\hypersetup{
    colorlinks, 
    linkcolor=blue, 
    citecolor=blue, 
    urlcolor=blue 
}

\newcommand{\user}[1]{\textbf{\textit{#1}}}

\begin{document}
\definecolor{R}{RGB}{255, 0, 0} 
\definecolor{B}{RGB}{0, 0, 255} 

\title{Multi-scale Temporal Fusion Transformer for Incomplete Vehicle Trajectory Prediction}

\author{Zhanwen Liu$^{1}$, Chao Li$^{1*}$, Yang Wang$^{1}$, Nan Yang$^{1}$, Xing Fan$^{2*}$, Jiaqi Ma$^{3}$, Xiangmo Zhao$^{1}$
\thanks{* Co-corresponding authors}
\thanks{$^{1}$Zhanwen Liu, Chao Li, Yang Wang, Nan Yang, Xiangmo Zhao are with the Department of Information Engineering, Chang’an University, Xi’an, Shaanxi 710018, PR China. {\tt\small zwliu@chd.edu.cn, lichao971204@foxmail.com, ywang120@chd.edu.cn, 2022024001@chd.edu.cn, xmzhao@chd.edu.cn}}
\thanks{$^{2}$Xing Fan is with the school of electronics and control engineering, Chang’an University, Xi’an, Shaanxi 710018, PR China. {\tt\small fanx@chd.edu.cn}}
\thanks{$^{3}$Jiaqi Ma is with the UCLA Mobility Lab and FHWA Center of Excellence on New Mobility and Automated Vehicles, University of California, Los Angeles (UCLA), CA 90095 USA. {\tt\small jiaqima@ucla.edu}}}

\markboth{Journal of \LaTeX\ Class Files,~Vol.~14, No.~8, August~2021}%
{Shell \MakeLowercase{\textit{et al.}}: A Sample Article Using IEEEtran.cls for IEEE Journals}

\IEEEpubid{0000--0000/00\$00.00~\copyright~2021 IEEE}

\maketitle

\begin{abstract}
Motion prediction plays an essential role in autonomous driving systems, enabling autonomous vehicles to achieve more accurate local-path planning and driving decisions based on predictions of the surrounding vehicles. However, existing methods neglect the potential missing values caused by object occlusion, perception failures, \textit{etc.}, which inevitably degrades the trajectory prediction performance in real traffic scenarios. To address this limitation, we propose a novel end-to-end framework for incomplete vehicle trajectory prediction, named Multi-scale Temporal Fusion Transformer (MTFT), which consists of the Multi-scale Attention Head (MAH) and the Continuity Representation-guided Multi-scale Fusion (CRMF) module. Specifically, the MAH leverages the multi-head attention mechanism to parallelly capture  multi-scale motion representation of trajectory from different temporal granularities, thus mitigating the adverse effect of missing values on prediction. Furthermore, the multi-scale motion representation is input into the CRMF module for multi-scale fusion to obtain the robust temporal feature of the vehicle. During the fusion process, the continuity representation of vehicle motion is first extracted across time steps to guide the fusion, ensuring that the resulting temporal feature incorporates both detailed information and the overall trend of vehicle motion, which facilitates the accurate decoding of future trajectory that is consistent with the vehicle’s motion trend. We evaluate the proposed model on four datasets derived from highway and urban traffic scenarios. The experimental results demonstrate its superior performance in the incomplete vehicle trajectory prediction task compared with state-of-the-art models, \textit{e.g.}, a comprehensive performance improvement of more than 39\% on the HighD dataset. 
\end{abstract}

\begin{IEEEkeywords}
Autonomous Driving, Motion Prediction, Incomplete Trajectory Prediction, Multi-scale Fusion.
\end{IEEEkeywords}

\section{Introduction}
\IEEEPARstart{W}{ith} the development of autonomous driving technologies, autonomous vehicles (AVs) have made great achievements. To guarantee the safety and efficiency in complex and dynamic traffic scenarios, AVs need to utilize the vehicle position and road topology information provided by roadside or onboard sensing systems to predict the future trajectories of surrounding vehicles \cite{ref1.7,ref1.8}, enhancing their own local-path planning and collision warning. 

In recent years, benefit from the development of deep learning technologies, various well-designed deep learning models have been proposed. Interaction-aware models are first proposed to consider inter-vehicle interaction in trajectory prediction, achieving promising prediction performance in simple highway scenarios \cite{ref4.5,ref1.1}. Subsequently, models leveraging high-definition (HD) maps have been introduced to incorporate the constraints imposed by complex road topologies, enabling flexible trajectory prediction of vehicles in non-Euclidean spaces \cite{ref1.2}. With the advancement of Transformer, numerous prediction models based on Transformers have recently emerged, harnessing attention mechanisms to capture long-term dependency and achieve superior performance in long-term trajectory prediction \cite{ref2.18}.

\begin{figure*}[thpb]
      \centering
      \includegraphics[scale=0.54]{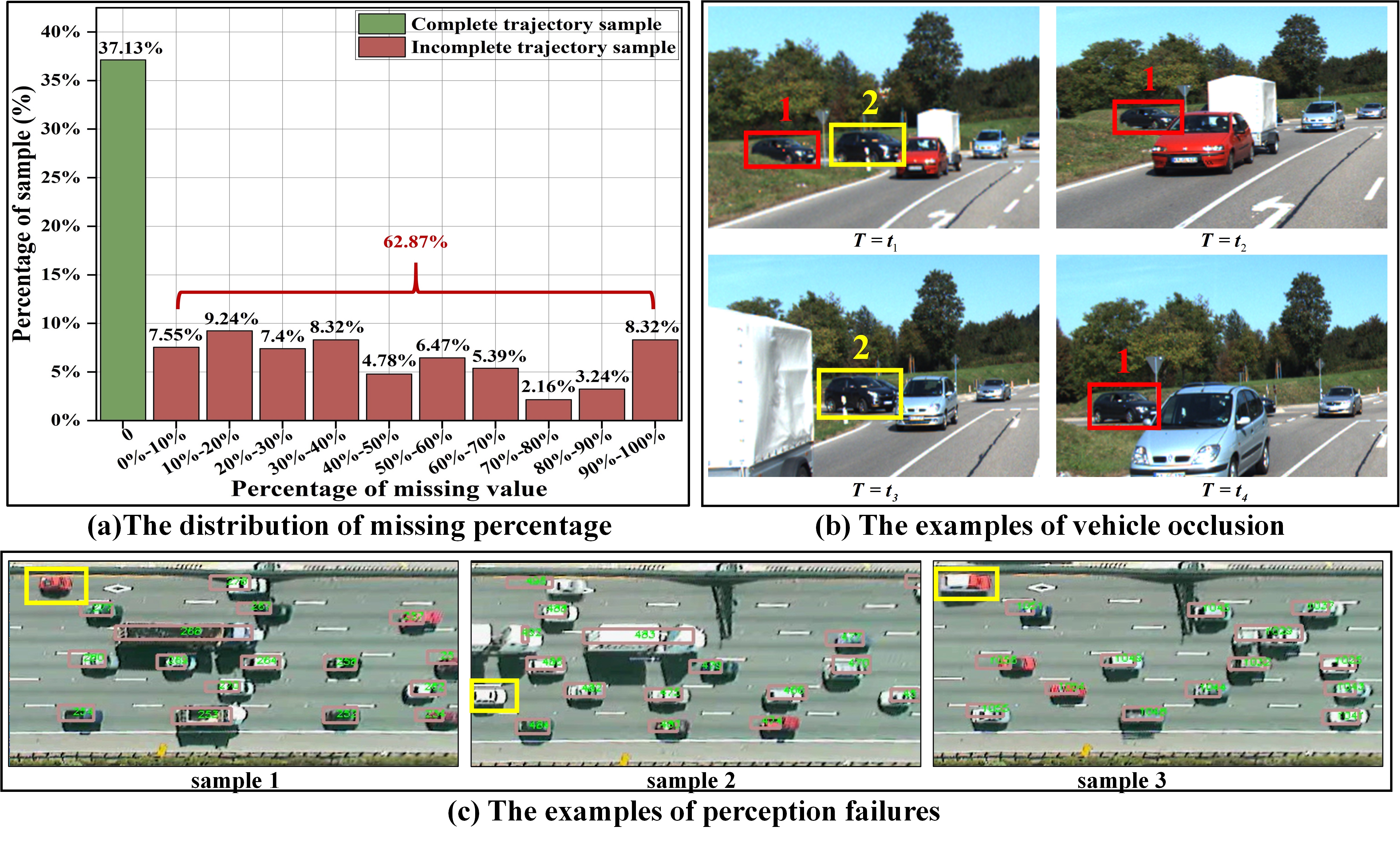}
      \caption{(a) lists the distribution of missing percentages of trajectory, revealing that most of the trajectory samples have varying percentages of missing values. In the case shown in (b), vehicle 1 is occluded at time ${t_3}$, while vehicle 2 is occluded at time ${t_2}$ and ${t_4}$, resulting in their incomplete trajectory. In contrast, the three cases given in (c) avoid the problem of vehicles occluding through the BEV perspective. However, the perception algorithm only captures the trajectory of most vehicles (marked by pink boxes), while some vehicles (marked by yellow boxes) are not captured due to the failure of the perception algorithm, which brings the incomplete trajectory.}
      \label{fig1.1}
\end{figure*}

While achieving impressive achievements, the performance of existing methods highly depended on the completeness of observed history trajectory, \textbf{\textit{neglecting the potential missing values}} caused by occlusion and perception failures in real traffic scenarios. To illustrate this issue intuitively, we analyze the vehicle object tracking dataset from KITTI [\cite{ref1.3}, considering vehicles labeled as “fully occluded” or “mostly occluded” as missing, and the statistical results depicted in  Fig. \ref{fig1.1} (a), which shows that a mere 37.13\% of the trajectory samples are complete, while 62.87\% of the trajectory samples contain missing values, and the missing percentages are randomly distributed in the interval (0\%, 100\%). The examples with missing values caused by object occlusion and perception failures are shown in Fig. \ref{fig1.1} (b) and Fig. \ref{fig1.1} (c), respectively. The existence of missing values hinders the model from capturing the accurate temporal dependence between the adjacent time steps, undoubtedly leading to a significant negative impact on the trajectory prediction performance of existing models.

\IEEEpubidadjcol 

To achieve end-to-end incomplete vehicle trajectory prediction,the \textbf{M}ulti-scale \textbf{T}emporal \textbf{F}usion \textbf{T}ransformer (MTFT) is presented to alleviate the negative impact of missing values on trajectory prediction by capturing and fusing the temporal dependency of trajectory across multiple time scales. Specifically, we design a novel \textbf{M}ulti-scale \textbf{A}ttention \textbf{H}ead (MAH) utilizing the inherent padding mask mechanism in the Transformer. With the predefined scale mask, the MAH can perceive the temporal dependency of incomplete trajectory from different time scales parallelly, enabling the extraction of multi-scale motion representation. Furthermore, a \textbf{C}ontinuity \textbf{R}epresentation-guided \textbf{M}ulti-scale \textbf{F}usion (CRMF) module is proposed, which computes the multi-scale continuity representation and uses it  to guide the multi-scale motion representation fusion, thereby extracting the robust temporal feature of the vehicle. The critical idea behind our method is that the observed trajectory at different time scales may skip some missing values, and predicting the current trajectory from different temporal granularities utilizing multi-scale motion representation can alleviate the negative impact of missing values. Furthermore, the continuity representation extracted across time steps reflects the overall trend of vehicle motion and is insensitive to the missing patterns of trajectory, albeit with a loss of some detail. It can guide the model to perform the fusion of multi-scale motion representation to extract temporal representation that balances detailed information with the overall motion trend to the fullest extent possible, which contributes to achieving accurate prediction of incomplete vehicle trajectory.

This work is extended on our previous conference version MSTF \cite{ref1.6}. Compared to MSTF, we extend our work in the following three aspects: 1) we explore the positive significance of the novel multi-scale fusion method, CRMF, on incomplete trajectory prediction; 2) we incorporate the impact of inter-vehicle interaction to enhance accurate incomplete trajectory prediction; 3) We validated the effectiveness of MTFT by comparing it with state-of-art models on three public datasets and one self-constructed dataset. In summary, the main contributions of our work are summarized below:

\begin{itemize}
\item We conduct a statistical analysis to reveal that the trajectory missing is inevitable in real traffic scenarios, which will disturb the accurate prediction of future vehicle trajectory. To address this issue, we develop an end-to-end framework named MTFT consisting of MAH and CRMF for incomplete trajectory prediction. The MAH is meticulously designed to capture the multi-scale dependency of trajectory from different temporal granularities, mitigating the negative impact of missing trajectory on vehicle trajectory prediction.
\item We propose a novel CRMF module, which adaptively extracts multi-scale continuity representation across time steps based on the missing pattern of trajectory. Utilizing the multi-scale continuity representation, CRMF fuses the multi-scale motion representation to obtain the temporal representation that both contains detailed information and the overall trend of motion, which facilitates the accurate incomplete trajectory prediction.
\item To validate the prediction performance of the proposed model on the real incomplete trajectory, we analyze and filter the trajectory of vehicle with the object type ‘OTHERS’ in the Argoverse dataset \cite{ref2.12}, constructing the Incomplete Argoverse (IArgoverse) dataset. The IAgoverse dataset includes incomplete vehicle trajectory resulting from occlusions \textit{etc}. in real traffic scenarios, providing a benchmark for the research in incomplete vehicle trajectory prediction.
\item Our method is comprehensively validated on four datasets spanning three different traffic scenarios. Both ablation experiments and comparative experiments indicate that the proposed MTFT outperforms state-of-the-art methods significantly, especially when the percentage of missing values is high.
\end{itemize}

\section{Related Work}
\subsection{Trajectory Prediction}
The objective of trajectory prediction is to enable autonomous vehicles to anticipate the future positions of surrounding vehicles, thereby facilitating efficient and safe driving in dynamic traffic environments. 

In previous research, physical models including single trajectory methods \cite{ref2.1}, Kalman filtering methods \cite{ref2.2}, and Monte Carlo methods \cite{ref2.3} have been proposed for vehicle trajectory prediction, which are computationally relatively simple but cannot be flexibly applied to dynamically changing traffic scenarios, such that most of these methods are only suitable for short-term prediction (no more than 1 s) \cite{ref2.4}. Unlike physics-based methods that use several physics models, machine learning-based methods apply data-driven models to predict trajectory, such as Gaussian Process \cite{ref2.5}, Support Vector Machine \cite{ref2.6}, Hidden Markov Model \cite{ref2.7}, Dynamic Bayesian Network \cite{ref2.8,ref2.29}, and so on. These methods utilize feature extraction from data to estimate the distribution of future trajectory, thus offering novel insights into trajectory prediction and driving the advancement of learning-based approaches. Recently, trajectory prediction methods based on deep learning have become increasingly popular due to their ability to achieve superior predictive performance in more complex scenarios by considering interaction-related factors. As a typical representative of RNN-based models, Social-LSTM \cite{ref2.9} innovatively embeds vehicle features by rasterizing traffic scenes for interaction extraction, and then sequentially decodes future trajectory through the recursive work mechanism of LSTM. Following this, Hyeon et al. \cite{ref2.10} utilize an encoder-decoder LSTM architecture. The LSTM encoder encodes the historical trajectory features, while the LSTM decoder employs beam search algorithm to solve for the K most likely future trajectories. Xing et al. \cite{ref2.11} use GMM to distinguish driving styles and utilize the LSTM followed by fully connected regression layers to predict trajectory based on driving styles. By calculating the distance between the vehicle and the centerline, Chang et al. \cite{ref2.12} propose an LSTM encoder-decoder baseline that takes the map information and social information into account and compares it with the Nearest Neighbor (NN) regression method. Considering lane information, Kawasaki et al. \cite{ref2.13} combine LSTM with KF for multi-modal trajectory prediction. Li et al. \cite{ref2.14} propose a Two-stream LSTM structure, an improvement based on LSTM, aiming to achieve superior performance in long-term trajectory prediction. Although some methods using RNN have achieved significant success in extracting features from Euclidean spatial data, they lack flexibility when applied to complex traffic scenarios with complex road topologies such as roundabouts and intersections. Therefore, graph-based methods have been proposed to adapt to the complexity of road topology, facilitating the vehicle trajectory prediction in non-Euclidean space. A hierarchical GNN named VectorNet \cite{ref2.15} encapsulates the sequential features of map elements and past trajectories with instance-wise subgraphs and models interaction with a global graph. MacFormer \cite{ref2.16} is the extension of VectorNet focusing on predicting rational destinations. With the significant achievements of Transformer \cite{ref2.17} in the field of NLP, Transformer-based models \cite{ref2.18,ref2.28} have been applied to this task to establish direct links for inputs via an attentional mechanism, enabling the models to capture long-term dependency of the trajectory for achieving more accurate prediction. 

However, these methods assume that vehicle observations are entirely complete, which is too strong an assumption to satisfy in practice. Existing methods are not applicable to the prediction of incomplete trajectory whose temporal dependency is disrupted by missing values.

\subsection{Sequence Imputation}
In response to the issue of missing sequential data across various domains, three categories of sequential imputation methods have been proposed, encompassing statistical imputation methods, machine learning imputation approaches, and deep learning imputation algorithms. 

The simple statistical imputation methods replace missing values with statistics or the most similar ones among the data. In contrast, traditional machine learning imputation solutions train a prediction model for missing value imputation, including tree-based imputation methods, regression-based imputation methods, compression-based imputation methods, and shallow neural network (SNN)-based imputation methods. The Tree-based imputation methods establish a decision tree model for each incomplete feature containing missing values, such as XGBI \cite{ref2.19}. The regression-based methods utilize linear regression models with multiple imputations to estimate missing values, \textit{e.g.}, IIM \cite{ref2.20}. Unlike other machine learning solutions, the compression-based methods construct a single prediction model for the whole incomplete dataset \cite{ref2.21}. Additionally, the SNN-based methods make use of a shallow neural network to impute missing values, such as RRSI \cite{ref2.22}. Moreover, deep learning models have been employed to address imputation problems, such as deep autoencoders (AEs) and generative adversarial networks (GANs). Hinton et al. \cite{ref2.23} initially proposed deep AE models. It has powerful density estimators that capture complex distributions. Building upon the foundation of deep AE models, various AE-based imputation algorithms are presented, including HI-VAE \cite{ref2.24} and MIWAE \cite{ref2.25}. On the other hand, the GAN model introduced by Goodfellow et al. \cite{ref2.26} builds an adversarial training framework between two players, namely the generator and the discriminator, to play a minimax game. The GAN model possesses powerful modeling capability to learn complex high-dimensional distributions \cite{ref2.27}.

Nevertheless, the existing imputation methods are not designed for the specific task of incomplete vehicle trajectory prediction \cite{ref1.4, ref1.5}, and the two-stage incomplete trajectory prediction framework of imputation followed by prediction brings extra parameters and computation burden, which hinders the lightweight and timeliness of autonomous driving systems. Therefore, we designed a novel framework called MTFT based on Transformer \cite{ref2.17}, which enables end-to-end incomplete trajectory prediction by extracting multi-scale motion representation and continuity representation.

\section{Methods}
\subsection{Problem Formulation}
To facilitate the study of the temporal evolution of trajectory, existing datasets have supplemented the vehicle trajectory that is missing due to target occlusion or perception failure through manual annotation, which results in the incomplete trajectory being unavailable. Therefore, we adopt the method of randomly masking to generate the incomplete trajectory in this work. 

Specifically, considering the complete observed trajectory of the target vehicle and its surrounding $N$ vehicles as $\textbf{\textit{X}} = \left\{ {{\textbf{\textit{x}}_{tar}},{\textbf{\textit{x}}_1},{\textbf{\textit{x}}_2}..,{\textbf{\textit{x}}_N}} \right\}$, and ${\textbf{\textit{x}}_i}$ can be further denoted as ${\textbf{\textit{x}}_i} = \left\{ {x_i^{t + 1},x_i^{t + 2},...,x_i^{t + {T_h}}} \right\}$. Where $x_i^t \in\mathbb{R}^2$ is the 2D coordinate of vehicle $i$ at time $t$, and ${T_h}$ represents the observation horizon. To simulate the incomplete trajectory caused by the object occlusion and perception failures in real traffic scenarios, we define the sequence mask $\bar{\textbf{\textit{M}}} = \left\{ {{{\bar{\textbf{\textit{m}}}}_{tar}},{{\bar{\textbf{\textit{m}}}}_1},{{\bar{\textbf{\textit{m}}}}_2}..,{{\bar{\textbf{\textit{m}}}}_N}} \right\}$ with the same dimension as \textbf{\textit{X}}, where ${\bar{\textbf{\textit{m}}}_i} = \left\{ {\bar m_i^{t + 1},\bar m_i^{t + 2},...,\bar m_i^{t + {T_h}}} \right\}$ is the sequence mask for vehicle $i$. The variable $\bar m_i^t$ is assigned a value of 0 if the observation of vehicle $i$ at time $t$ is missing, and 1 otherwise. The positions and number of missing observations are generated completely at random. Under this setting, the generated incomplete trajectory can be expressed as:
\begin{equation}
{\textbf{\textit{X}}_{in}} = \textbf{\textit{X}} \odot \bar{\textbf{\textit{M}}}
\end{equation}
where ${\textbf{\textit{X}}_{in}} = \left\{ {{\textbf{\textit{x}}_{tar}}_{in},{\textbf{\textit{x}}_{1,in}},{\textbf{\textit{x}}_{2,in}}..,{\textbf{\textit{x}}_{N,in}}} \right\}$ is the incomplete trajectory of the target vehicle and its surrounding $N$ vehicles after random masking, and $ \odot $ represents element-wise multiplication.

Based on the generated incomplete historical trajectory, this study aims to predict the future trajectory  $\hat{\textbf{\textit{y}}}=\left\{ {{{\hat y}^{t + {T_h} + 1}},{{\hat y}^{t + {T_h} + 2}},...,{{\hat y}^{t + {T_h} + {T_f}}}} \right\}$ of the target vehicle. Where ${T_f}$ is the prediction horizon, and ${\hat y^t}\in\mathbb{R}^2$ is predicted the 2D coordinate at time $t$.

\subsection{Model Framework}
\begin{figure*}[thpb]
      \centering
      \includegraphics[scale=0.41]{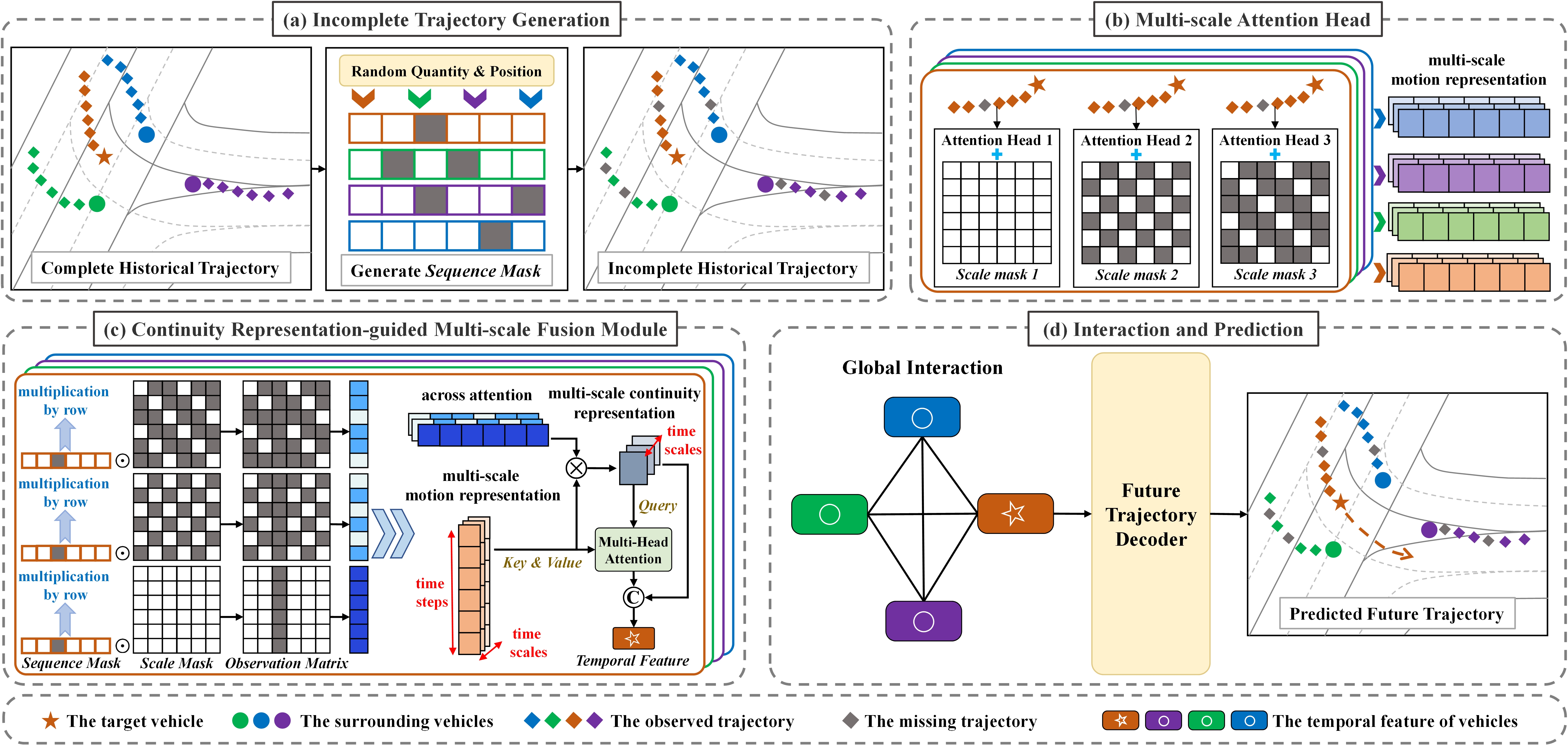}
      \caption{Illustration of the proposed MTFT framework. (a) Generate the sequence mask with randomly distributed number and position of masks, which is used to mask the complete trajectory provided by the public dataset to obtain incomplete trajectory. (b) Construct multi-scale attention head by predefined padding mask matrix with different temporal granularities for extracting multi-scale motion representation. (c) Extract multi-scale continuity representation across time steps and then use it as query vector for fusion of multi-scale motion representation. (d) Model the global interaction among all vehicles and output the predicted trajectory of the target vehicle.}
      \label{fig3.1}
\end{figure*}

Fig. \ref{fig3.1} depicts the pipeline of the proposed MTFT. Firstly, the sequence mask is obtained by randomly generating the number and distribution position of masks, which is applied to mask the complete trajectory sourced from the large public dataset to obtain the required incomplete trajectory. Secondly, utilizing the predefined scale mask, MAH captures the temporal dependency of incomplete trajectory from different time scales parallelly, thereby forming multi-scale motion representation. Subsequently, the CRMF module obtains across attention based on the observation matrix computed with the sequence mask and the scale mask, and the across attention is used to fuse multi-scale motion representation across time steps, yielding multi-scale continuity representation. Leveraging multi-scale continuity representation as query vector, CRMF fuses multi-scale motion representation across time scales to derive the temporal feature of each vehicle. Finally, following the modeling of global interactions among all vehicles, the future trajectory decoder outputs the predicted trajectory of the target vehicle.

\subsection{Multi-scale Attention Head}
Effectively capturing the temporal dependency of the trajectory is crucial for the vehicle trajectory prediction task. However, the presence of missing values disrupts the local dependency between the adjacent time steps. We argue that recurrent neural networks (\textit{e.g.}, LSTM and GRU), which extract local dependency of the trajectory serially using the recursive mechanism, are more susceptible to the negative effect of missing values. In contrast, the Transformer utilizes the attention mechanism to process the input sequence in parallel, enabling each value in the sequence to directly aggregate information from all the remaining values to capture global dependency, which mitigates the negative impact of certain missing values on prediction to a certain extent. Therefore, adopting a Transformer-designed encoder in our work is a natural decision.

Specifically, we first compute the query vector $\user{Q} = \left\{ {{\user{q}^1},{\user{q}^2},...,{\user{q}^n}} \right\}$, the key vector $\user{K} = \left\{ {{\user{k}^1},{\user{k}^2},...,{\user{k}^n}} \right\}$, and the value vector $\user{V} = \left\{ {{\user{v}^1},{\user{v}^2},...,{\user{v}^n}} \right\}$ for 
$n$ attention heads based on the incomplete trajectory ${\textbf{\textit{x}}_{in}}$.
\begin{equation}
\begin{array}{l}
\user{x}_{in}^{enc} = \beta ({\user{x}_{in}}) + Pos\\
{\user{q}^i} = {\varphi _Q}\left( {\user{x}_{in}^{enc},\user{W}_Q^i} \right)\\
{\user{k}^i} = {\varphi _K}\left( {\user{x}_{in}^{enc},\user{W}_K^i} \right)\\
{\user{v}^i} = {\varphi _V}\left( {\user{x}_{in}^{enc},\user{W}_V^i} \right)
\end{array}
\end{equation}
where $\beta $ denotes MLP, utilized to project the two-dimensional coordinate into higher dimensions, enhancing feature representation. Following the Transformer, positional encoding $Pos$ is incorporated into the model to distinguish the order of the input sequence. $\user{W}_Q^i$, $\user{W}_K^i$, and $\user{W}_V^i$ are the learnable parameter matrices for corresponding transformations ${\varphi _Q}$, ${\varphi _K}$, and ${\varphi _V}$, respectively.

Furthermore, the scale mask $\hat{\textbf{\textit{M}}}=\left\{ {{{\hat{\textbf{\textit{m}}}}^1},{{\hat{\textbf{\textit{m}}}}^2},...,{{\hat{\textbf{\textit{m}}}}^n}} \right\}$ with different temporal granularities is predefined for $n$ attention heads. Where, $\hat{{\textbf{\textit{m}}}}^i \in {\mathbb{R}^{len \times len}}$ is the scale mask for the attention head $i$, $len$ represents the length of the input sequence, and the value $\hat{\textbf{\textit{m}}}_{a,b}^i$ at position $\left( {a,b} \right)$ in $\hat{\textbf{\textit{m}}}^i$ can be expressed as:
\begin{equation}
\hat{\textbf{\textit{m}}}_{a,b}^i = \left\{ \begin{array}{l}
1,{\rm{ }}\frac{{a - b}}{i} \in \mathbb{Z}\\
0,{\rm{ }}Others
\end{array} \right.{\rm{     }}a,b \in \left\{ {1,2,...,len} \right\}
\end{equation}
where $\mathbb{Z}$ represents the set of integers.

\begin{figure}[thpb]
      \centering
      \includegraphics[scale=0.45]{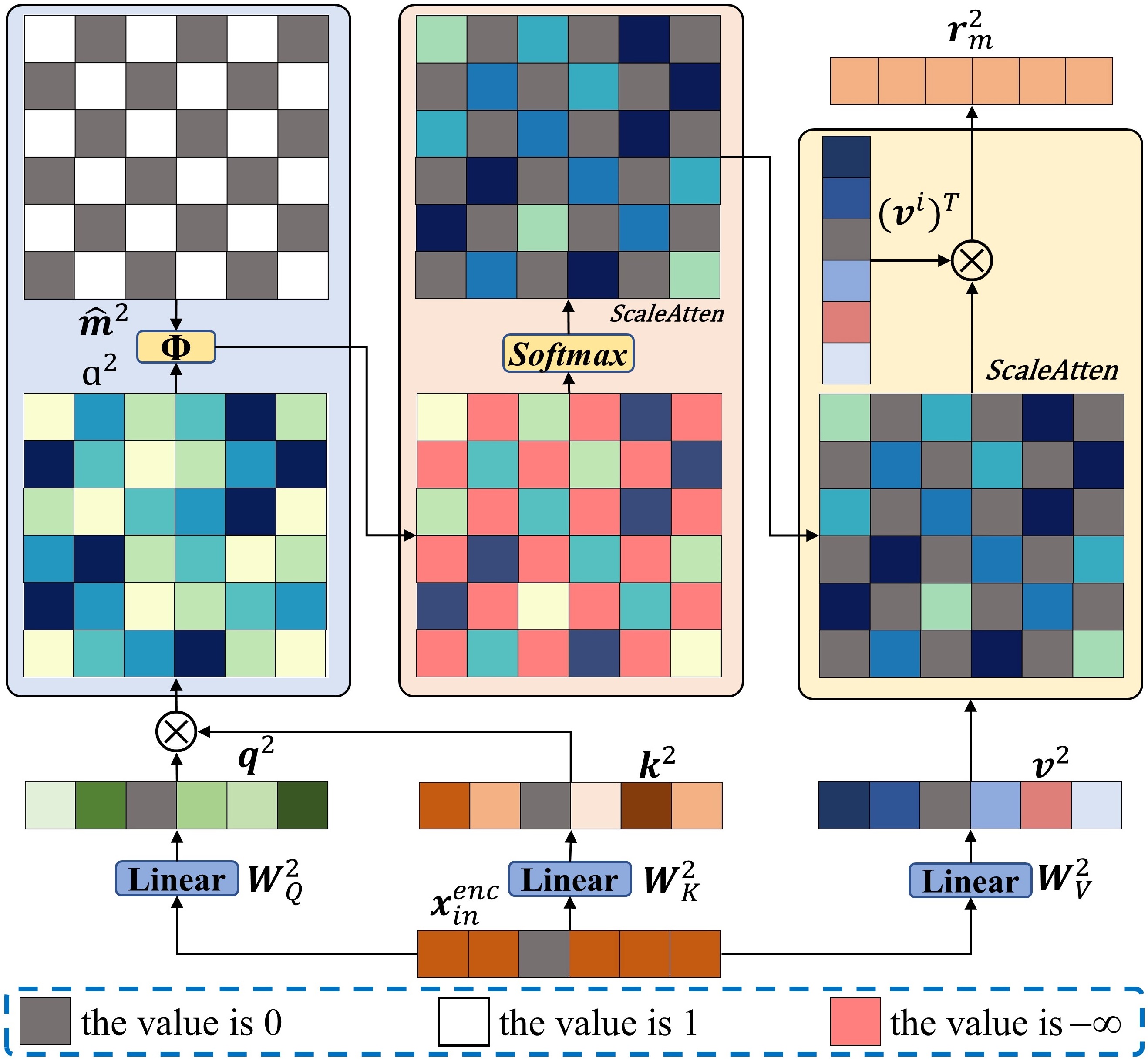}
      \caption{The computation process for attention head 2, where ${\hat{\user{m}}^2}$ the temporal scale of this attention head. The three special values 0, 1 and negative infinity in Formula (5) are represented by the gray, white and pink squares in this figure, respectively.}
      \label{fig3.2}
\end{figure}

Finally, leveraging the scale mask $\hat{\textbf{\textit{M}}}=\left\{ {{{\hat{\textbf{\textit{m}}}}^1},{{\hat{\textbf{\textit{m}}}}^2},...,{{\hat{\textbf{\textit{m}}}}^n}} \right\}$, we design the mapping function $\Phi$ to obtain the $ScaleAtten$, which enables the $n$ attention heads to observe the incomplete trajectory from different temporal granularities, parallelly extracting multi-scale motion representation ${\user{R}_m} = \left\{ {\user{r}_m^1,\user{r}_m^2,...,\user{r}_m^n} \right\}$, 
\begin{equation}
\begin{array}{l}
{\mit{\bm{\alpha}}} = {\user{q}^i}{\left( {{\user{k}^i}} \right)^T}\\
ScaleAtten\left( {{{\mit{\bm{\alpha}}}^i},{{\hat{\user{m}}}^i}} \right) = soft\max \left( {\frac{{\Phi \left( {{{\mit{\bm{\alpha}}}^i},{{\hat{\user{m}}}^i}} \right)}}{{\sqrt {d_k^i} }}} \right) \\
\user{r}_m^i = ScaleAtten\left( {{\mit{\bm{\alpha}}^i},{{\hat{\user{m}}}^i}} \right)*{\left( {{\user{v}^i}} \right)^T}
\end{array}
\end{equation}
where $\user{r}_m^i$ is the motion representation extracted by attention head $i$, and $d_k^i$ represents the dimension of the key vector ${k^i}$. The mapping rule of $\Phi$ is:
\begin{equation}
\Phi \left( {\alpha _{a,b}^i,\hat{\textbf{\textit{m}}}_{a,b}^i} \right) = \left\{ \begin{array}{l}
\alpha _{a,b}^i,{\rm{ }}\hat{\textbf{\textit{m}}}_{a,b}^i = 1\\
 - \infty ,{\rm{ }}\hat{\textbf{\textit{m}}}_{a,b}^i = 0
\end{array} \right.
\end{equation}
where $\alpha _{a,b}^i$ and $\hat{\textbf{\textit{m}}}_{a,b}^i$ are the values at position $\left( {a,b} \right)$ in ${{\mit{\bm{\alpha}}}^i}$ and $\hat{\textbf{\textit{m}}}^i$, respectively. We visualize the complete computation of the attention head 2, as shown in Fig. \ref{fig3.2}.

\subsection{Continuity Representation-guided Multi-scale Fusion Module}
The presence of missing values causes the encoded feature of the same incomplete trajectory sample to vary randomly with the missing pattern (the number and distributed positions of missing values), which poses a significant challenge to accurately decode the future trajectory. In this regard, we propose the continuity representation-guided multi-scale fusion (CRMF) module to extract continuity representation across time steps to guide the multi-scale representation fusion. The continuity representation does not contain detailed motion information but can reflect the overall trend of vehicle motion, and is insensitive to the trajectory missing. Guided by the continuity representation, the multi-scale motion representation fusion can obtain the temporal feature that both contains detailed motion information and reflects the overall trend of the motion to the greatest extent, which can enhance the prediction ability of the model.

Formally, the observation matrix $\user{S} = \left\{ {{\user{s}^1},{\user{s}^2},...,{\user{s}^n}} \right\}$ is first computed based on the randomly generated sequence mask $\bar{\user{m}}$ and the predefined scale mask $\hat{\textbf{\textit{M}}}=\left\{ {{{\hat{\textbf{\textit{m}}}}^1},{{\hat{\textbf{\textit{m}}}}^2},...,{{\hat{\textbf{\textit{m}}}}^n}} \right\}$. The process of computing the observation matrix can be formulated as follows:
\begin{equation}
{\user{s}^i} = \Lambda \left( {\bar{\user{m}},{{\hat{\user{m}}}^i}} \right)
\end{equation}
where the sequence mask $\bar{\user{m}}$ reflects the missing pattern of the trajectory, while the scale mask ${\hat{\user{m}}}^i$ serves as the observation scale of attention head $i$. ${\user{s}^i} \in {\mathbb{R}^{len \times len}}$ denotes the observation matrix at time scale $i$, which indicates whether the values in the sequence can be observed with each other under the constraints of the missing pattern and the time scale. The symbol $\Lambda$ represents the element-wise multiplication of $\bar{\user{m}}$ and ${\hat{\user{m}}}^i$  row by row.

Subsequently, utilizing the observation matrix ${\user{s}^i}$, the information increment ${\mit{\bm{\Delta}}} = {\left[ {\Delta _1^i,\Delta _2^i,...,\Delta _{len}^i} \right]^T}$ of each value in the sequence is statistically analyzed at time scale $i$.
\begin{equation}
\begin{array}{l}
s_{j,l}^i \in \left\{ {0,1} \right\}\\
\Delta _j^i = \sum\limits_{l = 1}^{len} {s_{j,l}^i} 
\end{array}
\end{equation}
where $s_{j,l}^i$ represents the value of the observation matrix ${\user{s}^i}$ at row $j$ and column $l$. $s_{j,l}^i$ is assigned 1 if the $l - th$ value in the sequence can be observed by the $j - th$ value, and 0 otherwise. $\Delta _j^i$ represents the information increment of the $j - th$ value in the sequence at time scale $i$.

Furthermore, due to the capability of MAH to capture global dependency, the feature of each trajectory point in multi-scale motion representation can reflect the motion continuity to some extent, only that the trajectory points at different locations observe the motion continuity from different perspectives. Therefore, we aggregate multi-scale motion representation across time steps, which enables CRMF to synthesize different perspectives to obtain robust continuity representation. Specifically, considering that the missing values affect the trajectory points at different locations differently, we compute the   according to the information increment ${{\mit{\bm{\Delta }}}^i} = {\left[ {\Delta _1^i,\Delta _2^i,...,\Delta _{len}^i} \right]^T}$, assigning greater weight to the trajectory points that are less affected by the missing values, and then deriving the multi-scale continuity representation ${\user{R}_c} = \left\{ {\user{r}_c^1,\user{r}_c^2,...,\user{r}_c^n} \right\}$ by weighted aggregation across time steps.

\begin{equation}
w_k^i = \frac{{\exp \left( {\Delta _k^i} \right)}}{{\sum\nolimits_{l = 1}^{len} {\exp \left( {\Delta _l^i} \right)} }},
\end{equation}
\begin{equation}
AcrossAtten\left( {{{\mit{\bm{\Delta }}}^i}} \right) = \left\{ {w_1^i,w_2^i,...,w_{len}^i} \right\},
\end{equation}
\begin{equation}
\user{r}_c^i = AcrossAtten\left( {{{\mit{\bm{\Delta }}}^i}} \right) \times {\left( {\user{r}_m^i} \right)^T},
\end{equation}
where $\user{r}_c^i$ is the continuity representation at time scale $i$.

Finally, to comprehensively capture the temporal feature ${\user{e}_{temp}}$ of the trajectory from multiple time scales based on the understanding of the overall continuity of the motion, we employ the multi-scale continuity representation ${\user{R}_c}$ as the query vector to fuse the multi-scale motion representation ${\user{R}_m}$.
\begin{equation}
{\user{Q}_c} = {\eta _{_Q}}\left( {{\user{R}_c},\user{W}_c^Q} \right)\\
\end{equation}
\begin{equation}
{\user{K}_m} = {\eta _K}\left( {{\user{R}_m},\user{W}_m^K} \right)\\
\end{equation}
\begin{equation}
{\user{V}_m} = {\eta _V}\left( {{\user{R}_m},\user{W}_m^V} \right)\\
\end{equation}
\begin{equation}
{\user{e}^{temp}} = {\mathop{\rm softmax}\nolimits} \left( {\frac{{{\user{Q}_c}{{\left( {{\user{K}_m}} \right)}^T}}}{{{d_K}}}} \right){\user{V}_m}
\end{equation}
where ${\eta _Q}$, ${\eta _K}$ and ${\eta _V}$ are transformation functions, which are achieved through MLP in our work. $\user{W}_c^Q$, $\user{W}_m^K$ and $\user{W}_m^V$ are their corresponding learnable parameters. The temporal feature ${\user{e}_{temp}}$ contains both detailed information about motion and reflects the overall continuity of motion to the greatest extent possible, and can robustly represent the temporal motion of the vehicle based on the input incomplete trajectory.

\subsection{Interaction and Prediction}
As a part of the transportation system, the target vehicle has to interact with the surrounding vehicles to achieve collision avoidance and efficient passage. Therefore, the extraction of inter-vehicle interaction is essential for accurate trajectory prediction. In this work, we model the inter-vehicle interaction with reference to the Global Interaction Module in Vectornet \cite{ref2.15}. Ultimately, the future trajectory decoder outputs the predicted trajectory $\hat{\user{y}}=\left\{ {{{\hat y}^{t + {T_h} + 1}},{{\hat y}^{t + {T_h} + 2}},...,{{\hat y}^{t + {T_h} + {T_f}}}} \right\}$ of the target vehicle for the next ${T_f}$ time steps.

\begin{equation}
{e_{i,j}} = \left\langle {\mathcal{G}\left( {\user{e}_i^{temp}} \right),\mathcal{G}\left( {\user{e}_j^{temp}} \right)} \right\rangle,
\end{equation}
\begin{equation}
{\alpha _{i,j}} = \frac{{\exp \left( {{e_{i,j}}} \right)}}{{\sum\limits_{k \in \left\{ {tar,1,2,,,N} \right\}} {\exp \left( {{e_{i,k}}} \right)} }},
\end{equation}
\begin{equation}
{\user{v}_i} = \sigma \left( {\sum\nolimits_{k \in \left\{ {tar,1,2,,,N} \right\}} {{\alpha _{i,j}}W\user{e}_j^{temp}} } \right),
\end{equation}
\begin{equation}
\hat{\user{y}}=\mathcal{P}\left( {{\user{v}_{tar}}} \right),
\end{equation}
where $\user{e}_i^{temp}$ is the temporal feature of vehicle $i$, and ${\user{v}_i}$ represents its final encoding after interaction modeling. The symbol $\left\langle \cdot \right\rangle $ denotes the computation of the inner product of vectors. $\mathcal{P}$ is the future trajectory decoder, which is implemented using $LSTM$ in our work.


\section{Experiment}
\subsection{Datasets}
In highway traffic scenarios, the simplicity of road alignment and lower vehicle density contribute to faster vehicle speed, characterized primarily by simple driving behaviors such as acceleration, deceleration, and lane changing. Conversely, within urban traffic scenarios, due to the complex road topology, high vehicle density, and strong interaction, the vehicle speed is slower but there are complex driving behaviors such as turning left, turning right, and U-turn. Considering the aforementioned distinctions, we conducted experiments to validate our proposed method on datasets corresponding to highway traffic scenarios, namely NGSIM [\cite{ref4.1,ref4.2} and HighD \cite{ref4.3}, as well as datasets representing urban traffic scenarios, namely Argoverse \cite{ref2.12} and IArgoverse.

\textbf{HighD}: The HighD team uses a drone to collect vehicle trajectory from six different locations on the Germany highway from the aerial perspective. The complete HighD dataset is captured at 25 Hz for 1.65 h at each of the six locations, which is composed of 60 recordings over areas of 400~420 meters span, with a mileage of 45,000 km, and more than 110, 000 vehicles are contained.

\textbf{NGSIM}: The NGSIM dataset is from the Next Generation Simulation research project initiated by the US Federal Highway Administration. The dataset extracts trajectory at 10 Hz from the image captured by digital video cameras and contains the data of all vehicles passing through the US-101 and the I-80 highways in 45 min.

\textbf{Argoverse}: Argoverse is a motion prediction benchmark, comprising over 30K data derived from the onboard sensing system in urban traffic scenarios. Each scenario is a 5-second sequence sampled at 10 Hz, and the task is to predict the position of the vehicle in the next 3 seconds based on its historical trajectory over 2 seconds. In our work, we only use historical vehicle trajectory for prediction and do not use map data such as rasterized drivable area maps and ground height maps provided by the benchmark.

\begin{figure}[thpb]
      \centering
      \includegraphics[scale=0.45]{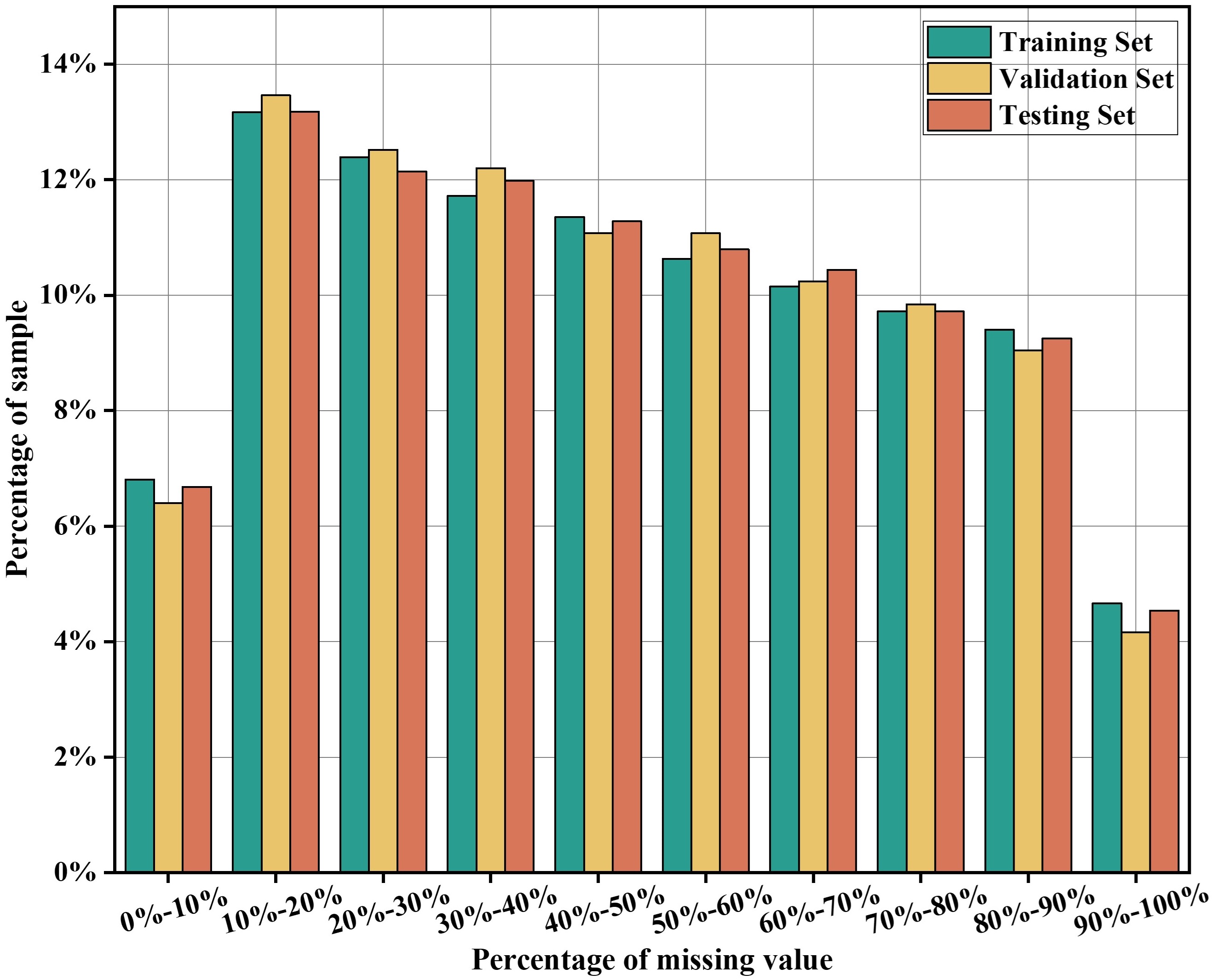}
      \caption{The distribution of the percentages of samples with different missing ratio in dataset IArgoverse.}
      \label{fig4.1}
\end{figure}

\textbf{IArgoverse}: In the analysis of the Argoverse dataset, it is found that the trajectories of ‘AGENT’ vehicles are complete, while those of 'OTHERS' vehicles contain missing values due to object occlusion and perception failures. Therefore, we filter the trajectories of ‘OTHERS’ vehicles to obtain the Incomplete Argoverse (IArgoverse) dataset, which is used to validate the prediction performance of the proposed model for real incomplete trajectory. Following Argoverse, we select only challenging prediction scenarios to compose the IArgoverse dataset. The selection criteria can be summarized as follows: (a) The missing proportion of 2-second history trajectory lies in the interval $\left( {0\%, 100\% } \right)$, which is used as an incomplete input, while the 3-second future trajectory is complete, serving as label in the model training or testing; (b) Within the 3-second prediction horizon, the vehicle exhibits lateral displacement exceeding 5 meters or longitudinal velocity changes exceeding 20 km/h, which is used as a challenging scenario to evaluate the model performance. Based on the above criteria, we obtain 105,233 samples for the training, 11,693 samples for validation, and 23,639 samples for the testing, respectively. The distribution of missing proportions for the IAgoverse dataset is illustrated in Fig. \ref{fig4.1}.

\subsection{Evaluation Metrics}
To facilitate performance comparisons, we follow previous work \cite{ref2.9,ref2.18,ref4.4,ref4.5,ref4.6,ref4.7} and use different evaluation metrics in the experiments based on highway traffic scenario datasets and urban traffic scenario datasets.

In the experiments based on the NGSIM and HighD datasets, we evaluate model performance at different prediction horizons using the root mean square error ($RMSE$). In contrast, average displacement error ($ADE$), final displacement error ($FDE$), and missing rate ($MR$) are adopted to comprehensively evaluate models in experiments based on the Argoverse and IArgoverse datasets, where $MR$ represents the ratio of samples whose $FDE$ is higher than 2 meters. It is worth noteworthy that our proposed model performs single prediction. To ensure a fair comparison, we also evaluate the performance of compared methods based on their single predictions, even though some methods give multiple possible predictions for the same sample.
\begin{equation}
RMSE = \sqrt {\frac{1}{m}\sum\limits_{i = 1}^m {\sum\limits_{t = {T_h} + 1}^{{T_h} + {T_f}} {{{\left( {\hat y_i^t - y_i^t} \right)}^2}} } },
\end{equation}
\begin{equation}
ADE = \frac{1}{m}\sum\limits_{i = 1}^m {\sum\limits_{t = {T_h} + 1}^{{T_h} + {T_f}} {\sqrt {{{\left( {\hat y_i^t - y_i^t} \right)}^2}} } },
\end{equation}
\begin{equation}
FDE = \frac{1}{m}\sum\limits_{i = 1}^m {\sqrt {{{\left( {\hat y_i^{{T_h} + {T_f}} - y_i^{{T_h} + {T_f}}} \right)}^2}} },
\end{equation}
where, $m$ is the number of samples. $\hat y_i^t$ and $y_i^t$ represents the predicted and true 2D coordinates of the sample $i$ at time $t$, respectively.

\subsection{Implementation Details}
The proposed MTFT comprises four layers, each incorporating five attention heads with different time scales, and all of them with a hidden layer dimension of 128. We implement the MTFT using PyTorch \cite{ref4.8} on 1 NVIDIA GeForce RTX 3090 with a batch size of 128, and the optimizer is Adam \cite{ref4.9}. To prevent overfitting, different learning rates and training epochs are set or different datasets with different data volumes.

\subsection{Ablation Experiments}
To validate the effectiveness of the proposed MAH and CRMF, we conduct ablation studies based on the four aforementioned datasets. Firstly, a baseline model is constructed utilizing the vanilla Transformer and interaction module. Subsequently, we incrementally add our proposed modules to the baseline to evaluate their impact on model performance. In the experiments, we consider the following models and evaluate their performance under the same parameter settings.

\textbf{V-TF}: This model comprises the vanilla Transformer and interaction module, serving as the baseline to demonstrate the effectiveness of MAH and CRMF.

\textbf{M-TF}: This model replaces the multi-head attention mechanism in V-TF with MAH, allowing it to capture multi-scale dependency of incomplete trajectory from different temporal granularities, which alleviates the negative impact of certain missing values on trajectory prediction. Comparing the performance of M-TF with that of V-TF demonstrates the effectiveness of MAH.

\textbf{MTFT}: This model integrates the CRMF module based on M-TF, enabling it to fuse multi-scale dependency guided by continuity representation. With the module, MTFT can extract temporal feature that not only contain detailed motion information but also maximally reflect motion continuity, thereby achieving more accurate prediction. Comparing the performance of MTFT with that of M-TF demonstrates the effectiveness of CRMF.

\begin{table*}[htbp]
\renewcommand{\tabcolsep}{11pt} 
\renewcommand{\arraystretch}{1.2} 
\centering
\caption{The results of ablation experiments based on HighD \& NGSIM datasets.The Best performance is marked in \textcolor{R}{\textbf{red bold}}.}
\label{table1}
\begin{tabular}{c|c|c|c|c|c|c|c|c|c|c|c}
\hline
\multicolumn{2}{c|}{Datasets} & \multicolumn{5}{c|}{HighD} & \multicolumn{5}{c}{NGSIM} \\
\cline{1-12}

\multicolumn{2}{c|}{Prediction horizons}&1 s&2 s&3 s&4 s&5 s&1 s&2 s&3 s&4 s&5 s \\
\cline{1-12}

\multirow{3}{*}{ $\left( {0\% ,30\% } \right]$}&V-TF&0.25&0.34&0.49&0.73&1.02&\textcolor{R}{\textbf{0.32}}&0.66&1.09&1.61&2.21 \\
\cline{2-12}
&M-TF&0.21&\textcolor{R}{\textbf{0.29}}&\textcolor{R}{\textbf{0.45}}&\textcolor{R}{\textbf{0.68}}&0.97&\textcolor{R}{\textbf{0.32}}&0.66&1.08&1.60&2.20  \\
\cline{2-12}
&MTFT&\textcolor{R}{\textbf{0.18}}&\textcolor{R}{\textbf{0.29}}&\textcolor{R}{\textbf{0.45}}&\textcolor{R}{\textbf{0.68}}&\textcolor{R}{\textbf{0.95}}&\textcolor{R}{\textbf{0.32}}&\textcolor{R}{\textbf{0.64}}&\textcolor{R}{\textbf{1.05}}&\textcolor{R}{\textbf{1.55}}&\textcolor{R}{\textbf{2.14}}\\ 
\hhline{=|=|=|=|=|=|=|=|=|=|=|=}

\multirow{3}{*}{ $\left( {30\% ,60\% } \right]$}&V-TF&0.27&0.35&0.51&0.75&1.05&\textcolor{R}{\textbf{0.33}}&0.68&1.11&1.64&2.24  \\
\cline{2-12}
&M-TF&0.23&0.31&0.48&0.71&1.01&\textcolor{R}{\textbf{0.33}}&0.68&1.11&1.63&2.23\\
\cline{2-12}
&MTFT&\textcolor{R}{\textbf{0.19}}&\textcolor{R}{\textbf{0.30}}&\textcolor{R}{\textbf{0.47}}&\textcolor{R}{\textbf{0.69}}&\textcolor{R}{\textbf{0.97}}&\textcolor{R}{\textbf{0.33}}&\textcolor{R}{\textbf{0.66}}&\textcolor{R}{\textbf{1.08}}&\textcolor{R}{\textbf{1.59}}&\textcolor{R}{\textbf{2.17}}\\ 
\hhline{=|=|=|=|=|=|=|=|=|=|=|=}

\multirow{3}{*}{ $\left( {60\% ,90\% } \right]$}&V-TF&0.31&0.40&0.59&0.84&1.17&0.36&0.72&1.18&1.71&2.33 \\
\cline{2-12}
&M-TF&0.27&0.37&0.55&0.82&1.15&\textcolor{R}{\textbf{0.35}}&0.72&1.16&1.70&2.31\\
\cline{2-12}
&MTFT&\textcolor{R}{\textbf{0.22}}&\textcolor{R}{\textbf{0.33}}&\textcolor{R}{\textbf{0.52}}&\textcolor{R}{\textbf{0.77}}&\textcolor{R}{\textbf{1.09}}&\textcolor{R}{\textbf{0.35}}&\textcolor{R}{\textbf{0.70}}&\textcolor{R}{\textbf{1.14}}&\textcolor{R}{\textbf{1.66}}&\textcolor{R}{\textbf{2.26}}\\ 

\hline

\end{tabular}
\end{table*}

\begin{table*}[htbp]
\renewcommand{\tabcolsep}{11pt} 
\renewcommand{\arraystretch}{1.2} 
\centering
\caption{The results of ablation experiments based on Argoverse \& IArgoverse datasets. The Best performance is marked in \textcolor{R}{\textbf{red bold}}.}
\label{table2}
\begin{tabular}{c|c|c|c|c||c|c|c||c|c|c}
\hline

\multirow{2}{*}{Datasets}&\multirow{2}{*}{Metrics}&\multicolumn{3}{c||}{$\left( {0\% ,30\% } \right]$}&\multicolumn{3}{c||}{$\left( {30\% ,60\% } \right]$}&\multicolumn{3}{c}{$\left( {60\% ,90\% } \right]$}\\
\cline{3-11} 

&&V-TF&M-TF&MTFT&V-TF&M-TF&MTFT&V-TF&M-TF&MTFT \\
\cline{1-11}

\multirow{3}{*}{Argoverse}&ADE&1.79&1.79&\textcolor{R}{\textbf{1.67}}&1.84&1.83&\textcolor{R}{\textbf{1.69}}&1.95&1.94&\textcolor{R}{\textbf{1.79}} \\
\cline{2-11}

&FDE&3.89&3.86&\textcolor{R}{\textbf{3.62}}&3.95&3.93&\textcolor{R}{\textbf{3.67}}&4.20&4.19&\textcolor{R}{\textbf{3.86}} \\
\cline{2-11}

&MR&0.64&0.63&\textcolor{R}{\textbf{0.60}}&0.66&0.64&\textcolor{R}{\textbf{0.60}}&0.68&0.67&\textcolor{R}{\textbf{0.62}} \\
\cline{1-11}

\multirow{5}{*}{IArgoverse}&ADE&1.16&1.15&\textcolor{R}{\textbf{1.14}}&1.96&1.91&\textcolor{R}{\textbf{1.86}}&2.93&2.80&\textcolor{R}{\textbf{2.74}} \\
\cline{2-11}

&FDE&1.83&1.79&\textcolor{R}{\textbf{1.75}}&3.60&3.46&\textcolor{R}{\textbf{3.37}}&6.02&5.53&\textcolor{R}{\textbf{5.37}} \\
\cline{2-11}

&MR&0.29&0.28&\textcolor{R}{\textbf{0.27}}&0.58&0.55&\textcolor{R}{\textbf{0.54}}&0.78&0.71&\textcolor{R}{\textbf{0.69}} \\
\cline{2-11}

&\multirow{2}{*}{Metrics}&\multicolumn{3}{c||}{1 s}&\multicolumn{3}{c||}{2 s}&\multicolumn{3}{c}{3 s} \\
\cline{3-11}

&&\multicolumn{9}{c}{Prediction horizons} \\
\hline

\end{tabular}
\end{table*}

\begin{figure}[thpb]
      \centering
      \includegraphics[scale=0.425]{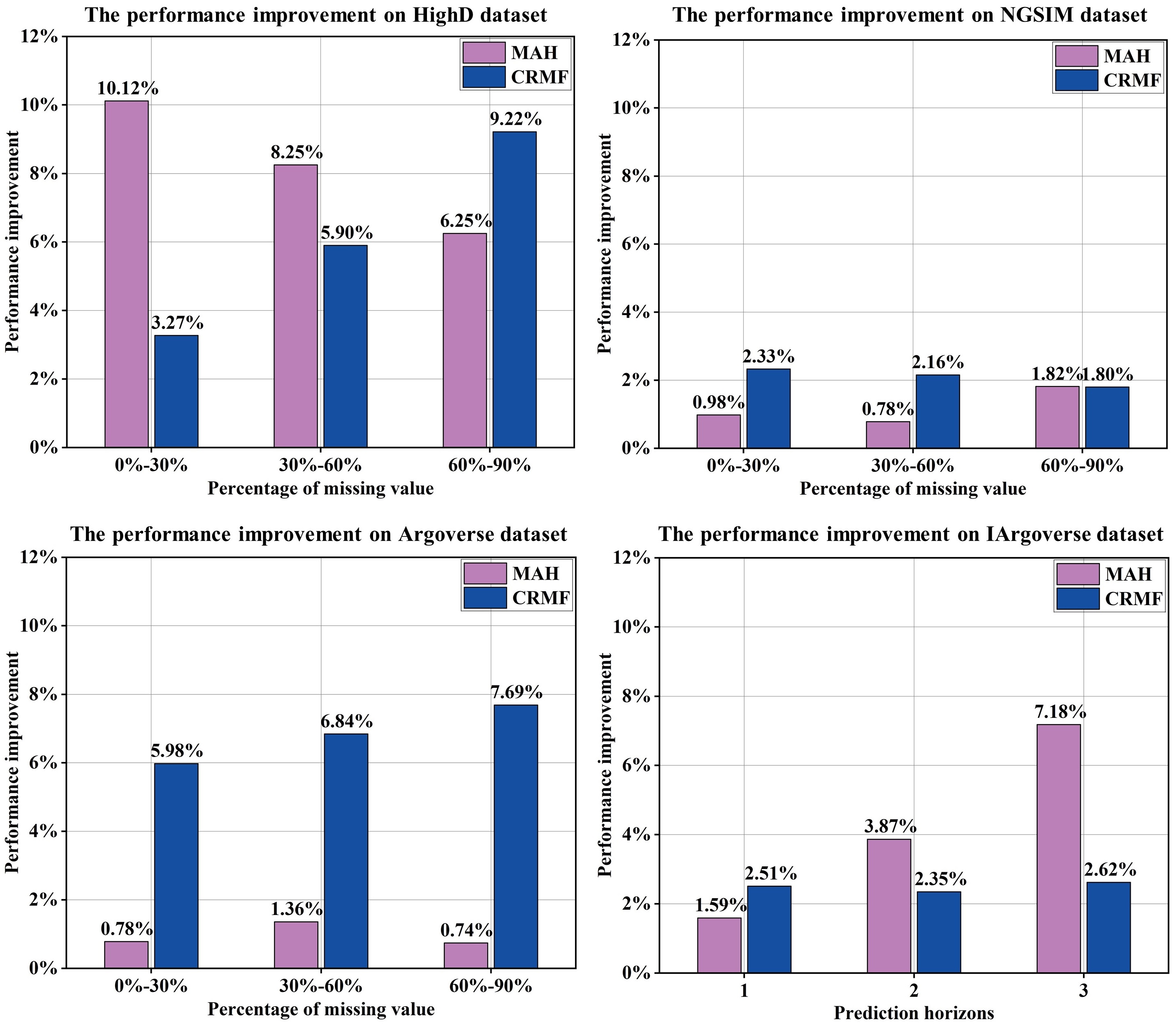}
      \caption{The performance improvement brought by the proposed MAH and CRMF modules on HighD, NGSIM, Argoverse, and IArgoverse datasets.}
      \label{fig4.2}
\end{figure}

The results of the ablation experiments conducted on the four aforementioned datasets are presented in Table \ref{table1} and Table \ref{table2}. To evaluate the prediction performance of these models under different missing percentages, we set three missing percentage intervals, which are $\left( {0\% ,30\% } \right]$, $\left( {30\% ,60\% } \right]$ and $\left( {60\% ,90\% } \right]$, respectively. Considering that the missing values in the IArgoverse dataset stem from real traffic scenarios rather than random masking, rendering the definition of specific missing intervals impractical. Therefore, we introduced prediction horizon as a variable in the experiment based on the IArgoverse dataset to comprehensively evaluate the model’s performance from multiple perspectives. Under different settings of missing intervals or prediction predictions, the detailed average performance improvements brought by MAH and CRMF are illustrated in  Fig. \ref{fig4.2}.

\begin{figure}[thpb]
      \centering
      \includegraphics[scale=0.37]{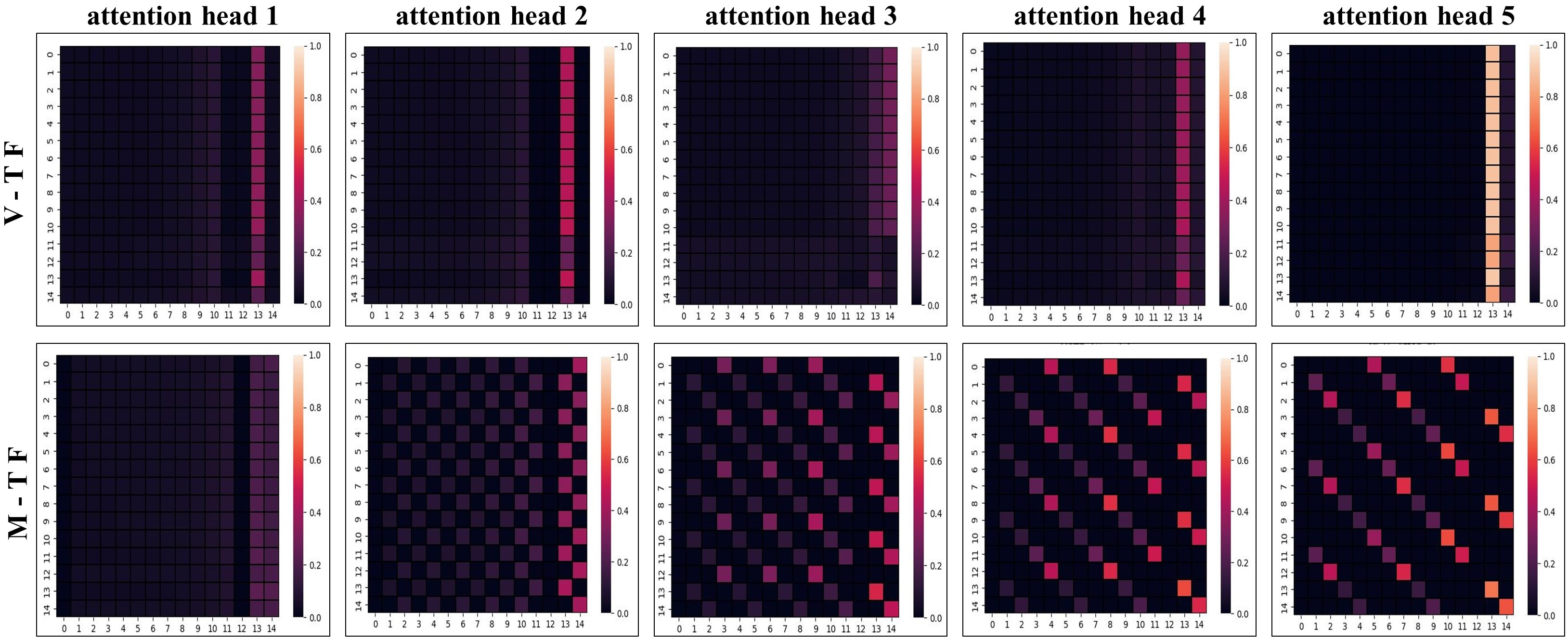}
      \caption{The visualization of attention matrices.}
      \label{fig4.3}
\end{figure}

\textit{1) MAH:} The comparison between M-TF and V-TF reveals that MAH brings a maximum performance improvement of over 10\% on the HighD dataset but less than 2\% on the NGSIM dataset under three different missing intervals. Influenced by the complex road topology and inter-vehicle interaction in urban traffic scenarios, the performance enhancement of MAH on the Argoverse and IArgoverse datasets ranges from 0.74\% to 7.18\%, which is lower than the average performance improvement observed on highway-based datasets. To visually demonstrate the effect of MAH, we visualize the attention matrices of five attention heads from V-TF and M-TF, as shown in Fig. \ref{fig4.3}. For any attention matrix, the attention value at the row $i$ and column $j$ represents the attention degree that the $i-th$ value in the sequence paid to the $j-th$ value during information aggregation. The visualization reveals that while V-TF can capture long-term dependency, it struggles to model multi-scale information autonomously, and all attention heads focus on historical information closer to the prediction moment. In contrast, M-TF’s different attention heads can focus on global dependencies across various scales, thereby capturing rich multi-scale motion representation, which can effectively mitigate the negative impact of missing trajectory on prediction. 

\textit{2) CRMF:} Comparison between MTFT and M-TF indicates that CRMF brings an average performance improvement of 6.13\% on the HighD dataset under three different missing interval settings, whereas the average improvement on the NGSIM dataset is only 2.10\%. For more complex urban traffic scenarios, the experiment based on the Argoverse dataset indicates that CRMF led to an average performance improvement of 6.81\%, while the improvement on the IArgoverse dataset was 2.49\%.

\textit{3) Summary:} Overall, both the multi-scale dependency captured by MAH and the multi-scale fusion performed by the CRMF module significantly enhance the model performance, validating the effectiveness of each module. Notably, the proposed modules achieve the highest average performance improvement on the HighD dataset, attributed to the simple traffic scenarios and high-quality annotated data in this dataset. Conversely, the presence of substantial data noise in the NGSIM dataset \cite{ref4.10} attenuates the positive effects of MAH and CRMF, resulting in the smallest performance improvement on the NGSIM dataset. Furthermore, the IArgoverse dataset exhibits more than 4\% of samples with a missing percentage exceeding 90\%, which makes the prediction performance of V-TF, M-TF and MTFT suboptimal on the IArgoverse dataset, even though the MAH and CRMF provide a significant performance improvement.

\subsection{Comparative Experiments}
To validate the superiority of the proposed MTFT, several state-of-the-art (SOTA) models from recent years are employed for comparitive experiments. Due to the necessity to evaluate the prediction performance of existing models for incomplete trajectory, only models with available code are selected for comparison in our work. In the experiments, the parameter settings of the following comparison models are maintained at their default values, and only the trajectory is randomly masked to generate incomplete input.

\begin{itemize}
\item \textbf{CS-LSTM} \cite{ref2.9}: This method integrates convolution operation into the social pooling layer, enabling the model to capture inter-vehicle interaction while preserving spatial information between vehicles. CS-LSTM outputs the two-dimensional Gaussian distribution of future trajectory points, achieving favorable prediction performance.prediction.
\item \textbf{PiP} \cite{ref4.4}: This model combines ego-vehicle trajectory planning with trajectory prediction of surrounding vehicles, generating predictions based on different planned trajectories of the ego-vehicle. PiP not only enhances prediction accuracy but also provides a highly integrated prediction-planning coupling process for autonomous driving systems.
\item \textbf{IH-Net} \cite{ref4.5}: This method extracts complete interaction information, including driver’s lane-changing intention and inter-vehicle interaction, through an intention-convolution social pooling module. Meanwhile, a novel hybrid attention mechanism is proposed to selectively reuse the historical information, which significantly improves the long-term prediction capability of the model.
\item \textbf{SSL} \cite{ref4.6}: This model incorporates self-supervised learning into the motion prediction task. Utilizing the pseudo-label freely available from the data, it designs four novel pretext tasks to learn superior and generalizable representations, markedly improving trajectory prediction performance in complex urban traffic scenarios.
\item \textbf{HLS} \cite{ref4.7}: This method proposes a hierarchical latent structure based on the VAE prediction model. It assumes that the trajectory distribution can be approximated as a mixture of simple distributions (or modes), representing each mode of the mixture and their weights by low-level and high-level latent variables, respectively, thereby achieving promising prediction performance.
\item \textbf{LAformer} \cite{ref2.18}: This model designs a temporally dense lane-aware estimation module to select high-potential lane segments from high-definition maps, which effectively and continuously aligns motion dynamics with scene information, leading to excellent generalized performance.
\item \textbf{MSTF} \cite{ref1.6}: This model is proposed in our previous research, which considers multi-scale dependency capture and continuity representation extraction in the incomplete trajectory prediction but neglects the design of the multi-scale fusion method and the modeling of inter-vehicle interaction.
\end{itemize}

As with the setup of the ablation experiments, we defined three missing intervals in the comparative experiments, which are $\left( {0\% ,30\% } \right]$, $\left( {30\% ,60\% } \right]$, and $\left( {60\% ,90\% } \right]$. The quantity and positions of missing values within these intervals are randomly generated.

Table \ref{table3} presents the comparative results based on the HighD and NGSIM datasets, which are derived from highway traffic scenarios. In general, the proposed model achieves optimal prediction accuracy across all experimental settings. From the perspective of prediction horizon, the average performance improvement of MTFT for 1 to 5-second prediction horizon on the HighD dataset is 39.27\%, 44.99\%, 45.54\%, 44.92\%, and 44.62\%, respectively; and that on the NGSIM dataset is 32.26\%, 31.93\%, 32.34\%, 33.09\%, and 33.90\% . The average performance enhancements of MTFT do not exhibit significant declines with increasing prediction horizons, which indicates that the performance advantage of MTFT is stable in long-term prediction as well. In terms of missing intervals, the average performance improvement of MTFT for the intervals $\left( {0\% ,30\% } \right]$, $\left( {30\% ,60\% } \right]$, and $\left( {60\% ,90\% } \right]$ on the HighD dataset is 39.67\%, 44.02\%, and 47.92\%, respectively; and that on the NGSIM dataset is 31.40\%, 32.54\%, and 34.17\%. The performance improvement of MTFT on both datasets increases with the higher missing percentage, indicating that our proposed model effectively mitigates the negative impact of trajectory incompleteness on prediction, and the larger the percentage of missing values, the more significant advantage MTFT exhibits in incomplete trajectory prediction. Moreover, PiP outperforms CS-LSTM in prediction accuracy on the complete trajectory but does not consistently surpass CS-LSTM in the incomplete trajectory prediction, which demonstrates that the existing models designed for complete trajectory prediction cannot flexibly transferable to the task of incomplete trajectory prediction in real traffic scenarios, underscoring the significance of our research.

\begin{table*}[htbp]
\renewcommand{\tabcolsep}{11pt} 
\renewcommand{\arraystretch}{1.2} 
\centering
\caption{The comparative results based on HighD \& NGSIM datasets.The Best performance is marked in \textcolor{R}{\textbf{red bold}}.}
\label{table3}
\begin{tabular}{c|c|c|c|c|c|c|c|c|c|c|c}
\hline
\multicolumn{2}{c|}{Datasets} & \multicolumn{5}{c|}{HighD} & \multicolumn{5}{c}{NGSIM} \\
\cline{1-12}

\multicolumn{2}{c|}{Prediction horizons}&1 s&2 s&3 s&4 s&5 s&1 s&2 s&3 s&4 s&5 s \\
\cline{1-12}

\multirow{4}{*}{ $\left( {0\% ,30\% } \right]$}&CS-LSTM&0.29&0.76&1.47&2.32&3.64&0.60&1.30&2.16&3.26&4.60 \\
\cline{2-12}

&PiP&0.54&1.21&2.10&3.22&4.58&0.61&1.29&2.11&3.13&4.38  \\
\cline{2-12}

&IH-Net&0.23&0.35&0.54&0.80&1.11&0.47&0.81&1.25&1.79&2.44\\ 
\cline{2-12}

&MSTF&0.19&0.30&0.47&0.69&0.96&\textcolor{R}{\textbf{0.32}}&\textcolor{R}{\textbf{0.64}}&1.12&1.73&2.44  \\
\cline{2-12}

&MTFT&\textcolor{R}{\textbf{0.18}}&\textcolor{R}{\textbf{0.29}}&\textcolor{R}{\textbf{0.45}}&\textcolor{R}{\textbf{0.68}}&\textcolor{R}{\textbf{0.95}}&\textcolor{R}{\textbf{0.32}}&\textcolor{R}{\textbf{0.64}}&\textcolor{R}{\textbf{1.05}}&\textcolor{R}{\textbf{1.55}}&\textcolor{R}{\textbf{2.14}} \\
\hhline{=|=|=|=|=|=|=|=|=|=|=|=}

\multirow{4}{*}{ $\left( {30\% ,60\% } \right]$}&CS-LSTM&0.31&0.81&1.52&2.40&3.72&0.61&1.33&2.21&3.33&4.70 \\
\cline{2-12}

&PiP&0.62&1.35&2.29&3.44&4.82&0.63&1.33&2.16&3.21&4.50  \\
\cline{2-12}

&IH-Net&0.26&0.40&0.60&0.87&1.20&0.51&0.88&1.36&1.94&2.58\\ 
\cline{2-12}

&MSTF&0.26&0.35&0.52&0.75&1.03&\textcolor{R}{\textbf{0.33}}&0.67&1.16&1.78&2.50  \\
\cline{2-12}

&MTFT&\textcolor{R}{\textbf{0.19}}&\textcolor{R}{\textbf{0.30}}&\textcolor{R}{\textbf{0.47}}&\textcolor{R}{\textbf{0.69}}&\textcolor{R}{\textbf{0.97}}&\textcolor{R}{\textbf{0.33}}&\textcolor{R}{\textbf{0.66}}&\textcolor{R}{\textbf{1.08}}&\textcolor{R}{\textbf{1.59}}&\textcolor{R}{\textbf{2.17}}  \\
\hhline{=|=|=|=|=|=|=|=|=|=|=|=}

\multirow{4}{*}{ $\left( {60\% ,90\% } \right]$}&CS-LSTM&0.39&0.97&1.72&2.65&3.96&0.66&1.42&2.37&3.59&5.06 \\
\cline{2-12}

&PiP&0.73&1.58&2.61&3.84&5.27&0.67&1.41&2.30&3.41&4.76  \\
\cline{2-12}

&IH-Net&0.31&0.47&0.71&1.04&1.45&0.56&0.98&1.51&2.13&2.86\\ 
\cline{2-12}

&MSTF&0.34&0.45&0.66&0.92&1.23&\textcolor{R}{\textbf{0.35}}&0.73&1.25&1.88&2.63  \\
\cline{2-12}

&MTFT&\textcolor{R}{\textbf{0.22}}&\textcolor{R}{\textbf{0.33}}&\textcolor{R}{\textbf{0.52}}&\textcolor{R}{\textbf{0.77}}&\textcolor{R}{\textbf{1.09}}&\textcolor{R}{\textbf{0.35}}&\textcolor{R}{\textbf{0.70}}&\textcolor{R}{\textbf{1.14}}&\textcolor{R}{\textbf{1.66}}&\textcolor{R}{\textbf{2.26}}   \\

\hline

\end{tabular}
\end{table*}

The quantitative comparative results based on the Argoverse dataset are summarized in Table \ref{table4}. When the missing percentage exceeds 30\%, the proposed MTFT achieves the average performance improvement of 20.96\%, 19.78\%, and 8.92\% in metrics ADE, FDE, and MR, while MTFT performs worse than SSL and LAformer when the missing percentage is below 30\%. We argue that the lane centerline in HD maps offers useful prior knowledge for trajectory prediction due to vehicle trajectory generally following the center of a lane \cite{ref2.12}, which enables SSL and LAformer to maintain higher prediction accuracy when the missing percentage is below 30\%. Nonetheless, when the missing interval increases to $\left( {30\% ,60\% } \right]$ or $\left( {60\% ,90\% } \right]$, the performance gain from lane priors is insufficient to offset the negative impact of trajectory missing. In contrast, our proposed MTFT achieves superior prediction performance without relying on prior information from HD maps by modeling the overall continuity of vehicle motion and integrating multi-scale dependency of incomplete trajectory.

\begin{table}[H]
\renewcommand{\tabcolsep}{5pt} 
\renewcommand{\arraystretch}{1.2}
\caption{The results of comparative experiment on Argoverse dataset.The Best performance is marked in \textcolor{R}{\textbf{red bold}}.}
\label{table4}
\centering
\begin{tabular}{c|c|c|c|c|c|c}
\hline
Missing Rate&Metric&SSL&HLS&LAformer&MSTF&MTFT\\
\cline{1-7}
\multirow{3}{*}{ $\left( {0\% ,30\% } \right]$}&ADE&1.57&2.20&\textcolor{R}{\textbf{1.54}}&1.91&1.67 \\
\cline{2-7}
                         &FDE&\textcolor{R}{\textbf{3.37}}&4.56&3.39&4.26&3.62 \\
\cline{2-7}                         
                         &MR&\textcolor{R}{\textbf{0.55}}&\textbackslash&0.56&0.68&0.60\\
\hhline{=|=|=|=|=|=|=}
\multirow{3}{*}{ $\left( {30\% ,60\% } \right]$}&ADE&1.86&2.40&1.86&1.95&\textcolor{R}{\textbf{1.69}} \\
\cline{2-7}
                         &FDE&3.91&4.96&4.10&4.34&\textcolor{R}{\textbf{3.67}} \\
\cline{2-7}                         
                         &MR&0.61&\textbackslash&0.63&0.69&\textcolor{R}{\textbf{0.60}}\\
\hhline{=|=|=|=|=|=|=}
\multirow{3}{*}{ $\left( {60\% ,90\% } \right]$}&ADE&2.67&2.85&2.27&2.11&\textcolor{R}{\textbf{1.79}} \\
\cline{2-7}
                         &FDE&5.41&5.66&4.97&4.67&\textcolor{R}{\textbf{3.86}} \\
\cline{2-7}                         
                         &MR&0.71&\textbackslash&0.70&0.72&\textcolor{R}{\textbf{0.62}} \\
\hline
\end{tabular}
\end{table}

\begin{figure*}[thpb]
      \centering
      \includegraphics[scale=0.37]{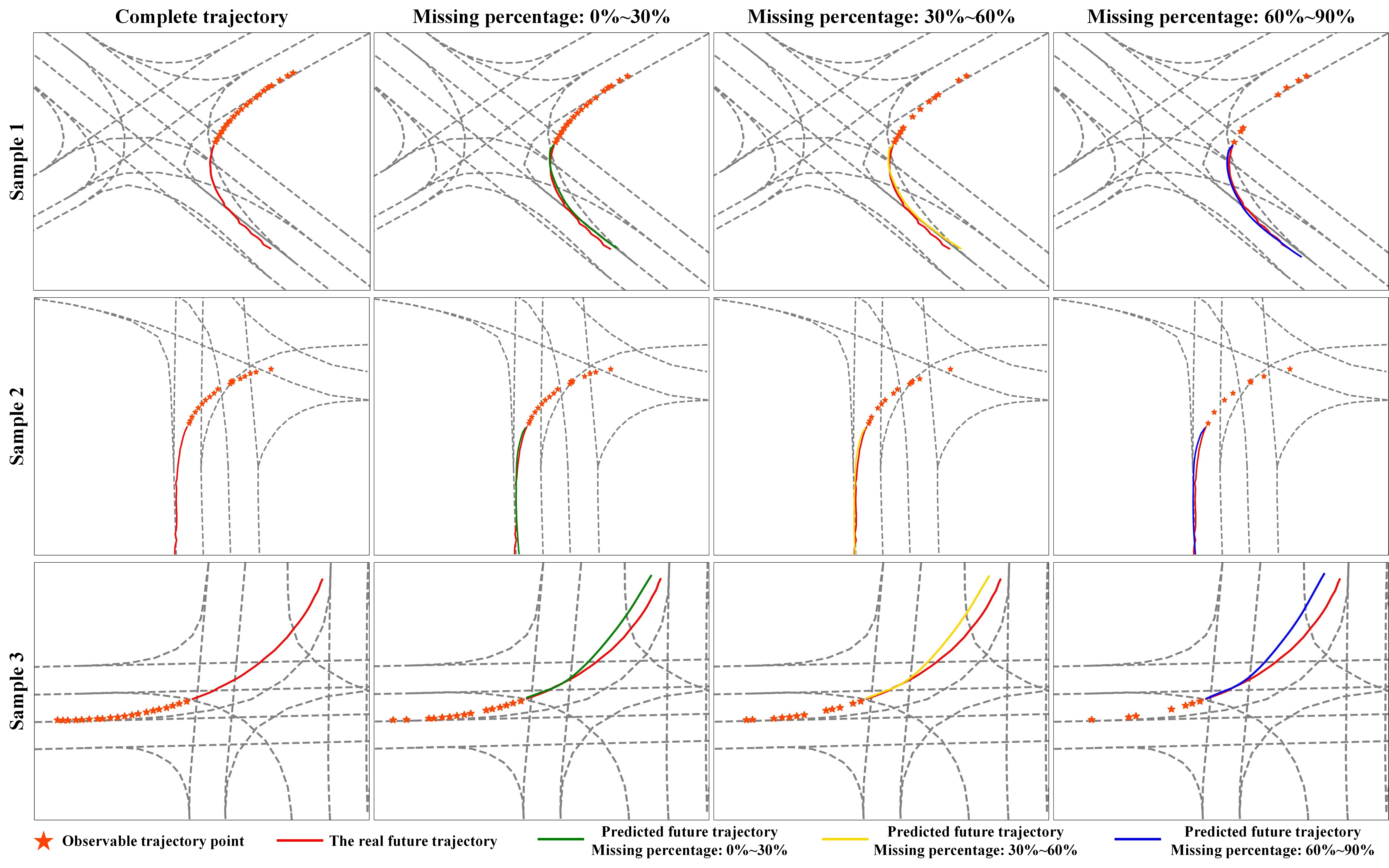}
      \caption{Visualization of predictions for three challenging scenarios at different missing percentage intervals.}
      \label{fig4.4}
\end{figure*}

To intuitively demonstrate the prediction performance of MTFT, we select three challenging scenarios from the Argoverse dataset and visualize their trajectory prediction results across three different missing intervals. As shown in Fig. \ref{fig4.4} , the prediction error does not exhibit a significant increase with the increasing missing percentage, which is attributed to the ability of MAH to capture multi-scale dependency from different temporal granularities, mitigating the negative impact of certain missing values on prediction, and thus enables the proposed MTFT to output more accurate predictions. Furthermore, without relying on HD maps as prior knowledge, the proposed MTFT leverages the CRMF module to extract and employ the high-level continuity of vehicle motion, resulting in the predicted trajectory consistent with overall motion trend and lane change.

\section{Conclusion}
This paper proposes a novel end-to-end framework named MTFT for incomplete vehicle trajectory prediction in real-world traffic scenarios, which comprises the Multi-scale Attention Head (MAH) and the Continuity Representation-guided Multi-scale Fusion (CRMF) module. The MAH is designed to extract multi-scale motion representation with global dependency from different time granularities, effectively alleviating the negative impact of missing values on prediction. Furthermore, the CRMF module can extract high-level continuity representation of vehicle motion and use it to guide the fusion of multi-scale motion representation. The temporal feature obtained after fusion not only contains detailed information about motion but also reflects the overall trend of vehicle motion to the greatest extent, which facilitates the accurate decoding of future trajectory that is consistent with the vehicle’s motion continuity. 

The future research direction is to explore the positive role of HD maps in the task of incomplete vehicle trajectory prediction. Since the vehicle trajectory generally follows the center of a lane, we attempt to use HD maps as prior knowledge to complement the missing information caused by missing trajectory points, enabling the model to output prediction consistent with the scene in complex urban traffic scenarios.

\section*{Acknowledgments}
This work was supported by the National Key Research and Development Program (No. 2023YFC3081700), the National Natural Science Foundation of China (No. 52172302) and the Two-chain Integration Key Special Project of Shaanxi Provincial Department of Science and Technology - Enterprise-Institute Joint Key special Project (2023-LL-QY-24).

\bibliography{IEEEref}

\begin{thebibliography}{10}
\providecommand{\url}[1]{#1}
\csname url@samestyle\endcsname
\providecommand{\newblock}{\relax}
\providecommand{\bibinfo}[2]{#2}
\providecommand{\BIBentrySTDinterwordspacing}{\spaceskip=0pt\relax}
\providecommand{\BIBentryALTinterwordstretchfactor}{4}
\providecommand{\BIBentryALTinterwordspacing}{\spaceskip=\fontdimen2\font plus
\BIBentryALTinterwordstretchfactor\fontdimen3\font minus \fontdimen4\font\relax}
\providecommand{\BIBforeignlanguage}[2]{{%
\expandafter\ifx\csname l@#1\endcsname\relax
\typeout{** WARNING: IEEEtran.bst: No hyphenation pattern has been}%
\typeout{** loaded for the language `#1'. Using the pattern for}%
\typeout{** the default language instead.}%
\else
\language=\csname l@#1\endcsname
\fi
#2}}
\providecommand{\BIBdecl}{\relax}
\BIBdecl

\bibitem{ref1.7}
Y.~H. Khalil and H.~T. Mouftah, ``Licanet: Further enhancement of joint perception and motion prediction based on multi-modal fusion,'' \emph{IEEE Open Journal of Intelligent Transportation Systems}, vol.~3, pp. 222--235, 2022.

\bibitem{ref1.8}
Y.~Tian, A.~Carballo, R.~Li, and K.~Takeda, ``Rsg-gcn: Predicting semantic relationships in urban traffic scene with map geometric prior,'' \emph{IEEE Open Journal of Intelligent Transportation Systems}, vol.~4, pp. 244--260, 2023.

\bibitem{ref4.5}
C.~Li, Z.~Liu, S.~Lin, Y.~Wang, and X.~Zhao, ``Intention-convolution and hybrid-attention network for vehicle trajectory prediction,'' \emph{Expert Systems with Applications}, vol. 236, p. 121412, 2024.

\bibitem{ref1.1}
C.~Li, Z.~Liu, N.~Yang, W.~Li, and X.~Zhao, ``Regional attention network with data-driven modal representation for multimodal trajectory prediction,'' \emph{Expert Systems with Applications}, vol. 232, p. 120808, 2023.

\bibitem{ref1.2}
Y.~Zhou, H.~Shao, L.~Wang, S.~L. Waslander, H.~Li, and Y.~Liu, ``Smartrefine: An scenario-adaptive refinement framework for efficient motion prediction,'' \emph{arXiv preprint arXiv:2403.11492}, 2024.

\bibitem{ref2.18}
M.~Liu, H.~Cheng, L.~Chen, H.~Broszio, J.~Li, R.~Zhao, M.~Sester, and M.~Y. Yang, ``Laformer: Trajectory prediction for autonomous driving with lane-aware scene constraints,'' \emph{arXiv preprint arXiv:2302.13933}, 2023.

\bibitem{ref1.3}
A.~Geiger, P.~Lenz, and R.~Urtasun, ``Are we ready for autonomous driving? the kitti vision benchmark suite,'' in \emph{2012 IEEE conference on computer vision and pattern recognition}.\hskip 1em plus 0.5em minus 0.4em\relax IEEE, 2012, pp. 3354--3361.

\bibitem{ref1.6}
Z.~Liu, C.~Li, N.~Yang, Y.~Wang, J.~Ma, G.~Cheng, and X.~Zhao, ``Mstf: Multiscale transformer for incomplete trajectory prediction,'' in \emph{2024 IEEE Intelligent Vehicles Symposium (IV)}, 2024, pp. 573--580.

\bibitem{ref2.12}
M.-F. Chang, J.~Lambert, P.~Sangkloy, J.~Singh, S.~Bak, A.~Hartnett, D.~Wang, P.~Carr, S.~Lucey, D.~Ramanan \emph{et~al.}, ``Argoverse: 3d tracking and forecasting with rich maps,'' in \emph{Proceedings of the IEEE/CVF conference on computer vision and pattern recognition}, 2019, pp. 8748--8757.

\bibitem{ref2.1}
M.~Br{\"a}nnstr{\"o}m, E.~Coelingh, and J.~Sj{\"o}berg, ``Model-based threat assessment for avoiding arbitrary vehicle collisions,'' \emph{IEEE Transactions on Intelligent Transportation Systems}, vol.~11, no.~3, pp. 658--669, 2010.

\bibitem{ref2.2}
V.~Lefkopoulos, M.~Menner, A.~Domahidi, and M.~N. Zeilinger, ``Interaction-aware motion prediction for autonomous driving: A multiple model kalman filtering scheme,'' \emph{IEEE Robotics and Automation Letters}, vol.~6, no.~1, pp. 80--87, 2020.

\bibitem{ref2.3}
Y.~Wang, Z.~Liu, Z.~Zuo, Z.~Li, L.~Wang, and X.~Luo, ``Trajectory planning and safety assessment of autonomous vehicles based on motion prediction and model predictive control,'' \emph{IEEE Transactions on Vehicular Technology}, vol.~68, no.~9, pp. 8546--8556, 2019.

\bibitem{ref2.4}
Y.~Huang, J.~Du, Z.~Yang, Z.~Zhou, L.~Zhang, and H.~Chen, ``A survey on trajectory-prediction methods for autonomous driving,'' \emph{IEEE Transactions on Intelligent Vehicles}, vol.~7, no.~3, pp. 652--674, 2022.

\bibitem{ref2.5}
Y.~Guo, V.~V. Kalidindi, M.~Arief, W.~Wang, J.~Zhu, H.~Peng, and D.~Zhao, ``Modeling multi-vehicle interaction scenarios using gaussian random field,'' in \emph{2019 IEEE Intelligent Transportation Systems Conference (ITSC)}.\hskip 1em plus 0.5em minus 0.4em\relax IEEE, 2019, pp. 3974--3980.

\bibitem{ref2.6}
P.~Kumar, M.~Perrollaz, S.~Lefevre, and C.~Laugier, ``Learning-based approach for online lane change intention prediction,'' in \emph{2013 IEEE Intelligent Vehicles Symposium (IV)}.\hskip 1em plus 0.5em minus 0.4em\relax IEEE, 2013, pp. 797--802.

\bibitem{ref2.7}
Y.~Wang, C.~Wang, W.~Zhao, and C.~Xu, ``Decision-making and planning method for autonomous vehicles based on motivation and risk assessment,'' \emph{IEEE Transactions on Vehicular Technology}, vol.~70, no.~1, pp. 107--120, 2021.

\bibitem{ref2.8}
Y.~Li, X.-Y. Lu, J.~Wang, and K.~Li, ``Pedestrian trajectory prediction combining probabilistic reasoning and sequence learning,'' \emph{IEEE transactions on intelligent vehicles}, vol.~5, no.~3, pp. 461--474, 2020.

\bibitem{ref2.29}
A.~Nayak, A.~Eskandarian, and Z.~Doerzaph, ``Uncertainty estimation of pedestrian future trajectory using bayesian approximation,'' \emph{IEEE Open Journal of Intelligent Transportation Systems}, vol.~3, pp. 617--630, 2022.

\bibitem{ref2.9}
N.~Deo and M.~M. Trivedi, ``Convolutional social pooling for vehicle trajectory prediction,'' in \emph{Proceedings of the IEEE conference on computer vision and pattern recognition workshops}, 2018, pp. 1468--1476.

\bibitem{ref2.10}
S.~H. Park, B.~Kim, C.~M. Kang, C.~C. Chung, and J.~W. Choi, ``Sequence-to-sequence prediction of vehicle trajectory via lstm encoder-decoder architecture,'' in \emph{2018 IEEE intelligent vehicles symposium (IV)}.\hskip 1em plus 0.5em minus 0.4em\relax IEEE, 2018, pp. 1672--1678.

\bibitem{ref2.11}
Y.~Xing, C.~Lv, and D.~Cao, ``Personalized vehicle trajectory prediction based on joint time-series modeling for connected vehicles,'' \emph{IEEE Transactions on Vehicular Technology}, vol.~69, no.~2, pp. 1341--1352, 2019.

\bibitem{ref2.13}
A.~Kawasaki and A.~Seki, ``Multimodal trajectory predictions for urban environments using geometric relationships between a vehicle and lanes,'' in \emph{2020 IEEE International Conference on Robotics and Automation (ICRA)}.\hskip 1em plus 0.5em minus 0.4em\relax IEEE, 2020, pp. 9203--9209.

\bibitem{ref2.14}
C.~Li, Z.~Liu, J.~Zhang, Y.~Wang, F.~Ding, and X.~Zhao, ``Two-stream lstm network with hybrid attention for vehicle trajectory prediction,'' in \emph{2022 IEEE 25th International Conference on Intelligent Transportation Systems (ITSC)}.\hskip 1em plus 0.5em minus 0.4em\relax IEEE, 2022, pp. 1927--1934.

\bibitem{ref2.15}
J.~Gao, C.~Sun, H.~Zhao, Y.~Shen, D.~Anguelov, C.~Li, and C.~Schmid, ``Vectornet: Encoding hd maps and agent dynamics from vectorized representation,'' in \emph{Proceedings of the IEEE/CVF Conference on Computer Vision and Pattern Recognition}, 2020, pp. 11\,525--11\,533.

\bibitem{ref2.16}
J.~Gu, C.~Sun, and H.~Zhao, ``Densetnt: End-to-end trajectory prediction from dense goal sets,'' in \emph{Proceedings of the IEEE/CVF International Conference on Computer Vision}, 2021, pp. 15\,303--15\,312.

\bibitem{ref2.17}
A.~Vaswani, N.~Shazeer, N.~Parmar, J.~Uszkoreit, L.~Jones, A.~N. Gomez, {\L}.~Kaiser, and I.~Polosukhin, ``Attention is all you need,'' \emph{Advances in neural information processing systems}, vol.~30, 2017.

\bibitem{ref2.28}
Z.~Wang, J.~Guo, Z.~Hu, H.~Zhang, J.~Zhang, and J.~Pu, ``Lane transformer: A high-efficiency trajectory prediction model,'' \emph{IEEE Open Journal of Intelligent Transportation Systems}, vol.~4, pp. 2--13, 2023.

\bibitem{ref2.19}
T.~Chen and C.~Guestrin, ``Xgboost: A scalable tree boosting system,'' in \emph{Proceedings of the 22nd acm sigkdd international conference on knowledge discovery and data mining}, 2016, pp. 785--794.

\bibitem{ref2.20}
A.~Zhang, S.~Song, Y.~Sun, and J.~Wang, ``Learning individual models for imputation,'' in \emph{2019 IEEE 35th International Conference on Data Engineering (ICDE)}.\hskip 1em plus 0.5em minus 0.4em\relax IEEE, 2019, pp. 160--171.

\bibitem{ref2.21}
J.~Josse, J.~Pag{\`e}s, and F.~Husson, ``Multiple imputation in principal component analysis,'' \emph{Advances in data analysis and classification}, vol.~5, pp. 231--246, 2011.

\bibitem{ref2.22}
B.~Muzellec, J.~Josse, C.~Boyer, and M.~Cuturi, ``Missing data imputation using optimal transport,'' in \emph{International Conference on Machine Learning}.\hskip 1em plus 0.5em minus 0.4em\relax PMLR, 2020, pp. 7130--7140.

\bibitem{ref2.23}
G.~E. Hinton and R.~R. Salakhutdinov, ``Reducing the dimensionality of data with neural networks,'' \emph{science}, vol. 313, no. 5786, pp. 504--507, 2006.

\bibitem{ref2.24}
A.~Nazabal, P.~M. Olmos, Z.~Ghahramani, and I.~Valera, ``Handling incomplete heterogeneous data using vaes,'' \emph{Pattern Recognition}, vol. 107, p. 107501, 2020.

\bibitem{ref2.25}
P.-A. Mattei and J.~Frellsen, ``Miwae: Deep generative modelling and imputation of incomplete data sets,'' in \emph{International conference on machine learning}.\hskip 1em plus 0.5em minus 0.4em\relax PMLR, 2019, pp. 4413--4423.

\bibitem{ref2.26}
I.~Goodfellow, J.~Pouget-Abadie, M.~Mirza, B.~Xu, D.~Warde-Farley, S.~Ozair, A.~Courville, and Y.~Bengio, ``Generative adversarial networks,'' \emph{Communications of the ACM}, vol.~63, no.~11, pp. 139--144, 2020.

\bibitem{ref2.27}
I.~Spinelli, S.~Scardapane, and A.~Uncini, ``Missing data imputation with adversarially-trained graph convolutional networks,'' \emph{Neural Networks}, vol. 129, pp. 249--260, 2020.

\bibitem{ref1.4}
M.~Qi, J.~Qin, Y.~Wu, and Y.~Yang, ``Imitative non-autoregressive modeling for trajectory forecasting and imputation,'' in \emph{Proceedings of the IEEE/CVF Conference on Computer Vision and Pattern Recognition}, 2020, pp. 12\,736--12\,745.

\bibitem{ref1.5}
Y.~Xu, A.~Bazarjani, H.-g. Chi, C.~Choi, and Y.~Fu, ``Uncovering the missing pattern: Unified framework towards trajectory imputation and prediction,'' in \emph{Proceedings of the IEEE/CVF Conference on Computer Vision and Pattern Recognition}, 2023, pp. 9632--9643.

\bibitem{ref4.1}
J.~Colyar and J.~Halkias, ``Us highway 80 dataset, federal highway administration (fhwa), vol,'' \emph{Tech, no. Rep}, 2006.

\bibitem{ref4.2}
------, ``Us highway 101 dataset. federal highway administration research and technology fact sheet. publication number: Fhwa-hrt-07-030,'' \emph{Tech. Rep.}, 2007.

\bibitem{ref4.3}
R.~Krajewski, J.~Bock, L.~Kloeker, and L.~Eckstein, ``The highd dataset: A drone dataset of naturalistic vehicle trajectories on german highways for validation of highly automated driving systems,'' in \emph{2018 21st international conference on intelligent transportation systems (ITSC)}.\hskip 1em plus 0.5em minus 0.4em\relax IEEE, 2018, pp. 2118--2125.

\bibitem{ref4.4}
H.~Song, W.~Ding, Y.~Chen, S.~Shen, M.~Y. Wang, and Q.~Chen, ``Pip: Planning-informed trajectory prediction for autonomous driving,'' in \emph{Computer Vision--ECCV 2020: 16th European Conference, Glasgow, UK, August 23--28, 2020, Proceedings, Part XXI 16}.\hskip 1em plus 0.5em minus 0.4em\relax Springer, 2020, pp. 598--614.

\bibitem{ref4.6}
P.~Bhattacharyya, C.~Huang, and K.~Czarnecki, ``Ssl-lanes: Self-supervised learning for motion forecasting in autonomous driving,'' in \emph{Conference on Robot Learning}.\hskip 1em plus 0.5em minus 0.4em\relax PMLR, 2023, pp. 1793--1805.

\bibitem{ref4.7}
D.~Choi and K.~Min, ``Hierarchical latent structure for multi-modal vehicle trajectory forecasting,'' in \emph{European Conference on Computer Vision}.\hskip 1em plus 0.5em minus 0.4em\relax Springer, 2022, pp. 129--145.

\bibitem{ref4.8}
A.~Paszke, S.~Gross, F.~Massa, A.~Lerer, J.~Bradbury, G.~Chanan, T.~Killeen, Z.~Lin, N.~Gimelshein, L.~Antiga \emph{et~al.}, ``Pytorch: An imperative style, high-performance deep learning library,'' \emph{Advances in neural information processing systems}, vol.~32, 2019.

\bibitem{ref4.9}
D.~P. Kingma and J.~Ba, ``Adam: A method for stochastic optimization,'' \emph{arXiv preprint arXiv:1412.6980}, 2014.

\bibitem{ref4.10}
B.~Coifman and L.~Li, ``A critical evaluation of the next generation simulation (ngsim) vehicle trajectory dataset,'' \emph{Transportation Research Part B: Methodological}, vol. 105, pp. 362--377, 2017.

\end{thebibliography}

\vspace{-20pt}

\begin{IEEEbiography}[{\includegraphics[width=1in,height=1.25in,clip,keepaspectratio]{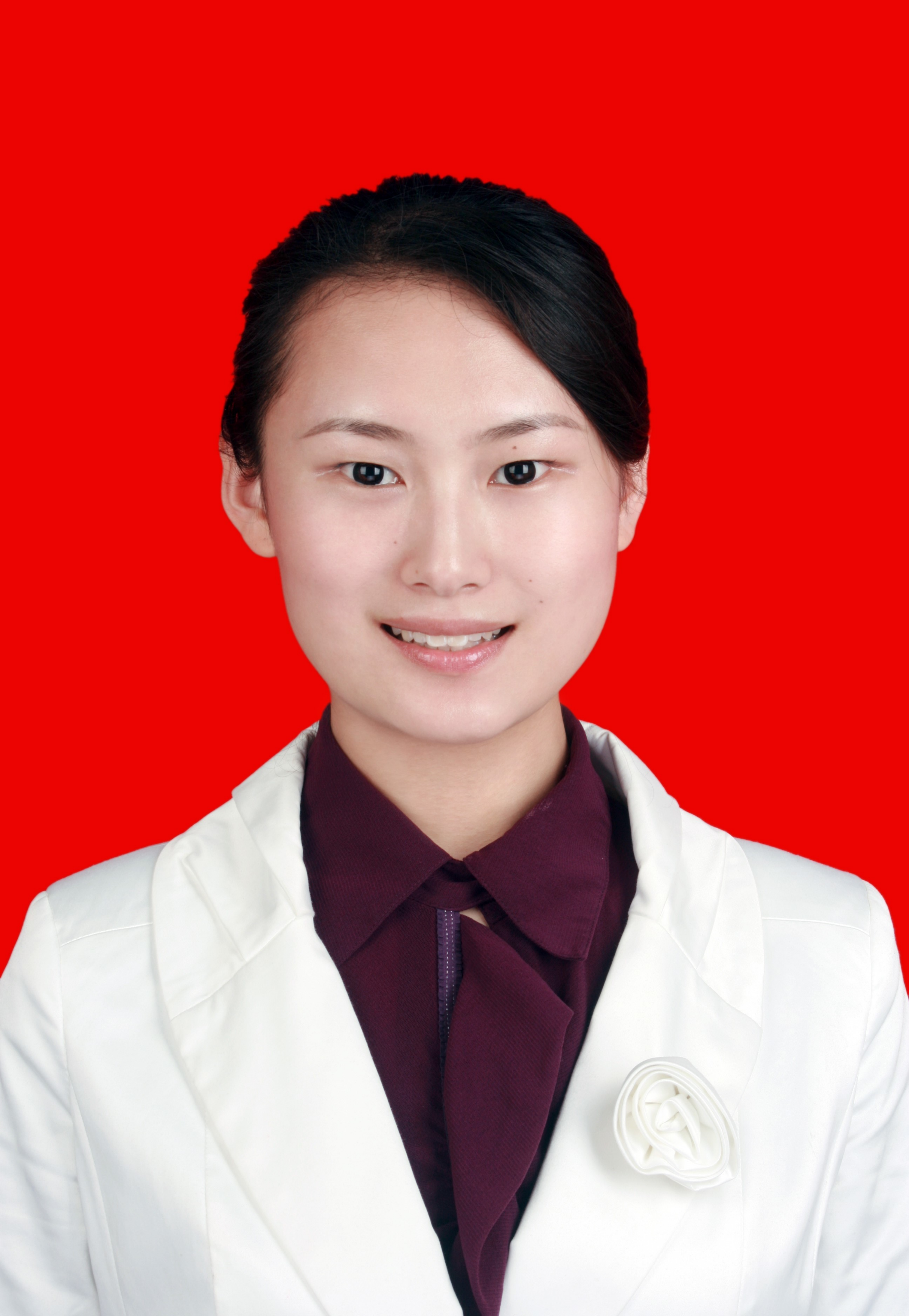}}]{Zhanwen Liu}
(Member, IEEE) received the B.S. degree from Northwestern Polytechnical University, Xi'an, China, in 2006, the M.S. and the Ph.D. degrees in Traffic Information Engineering and Control from Chang’an University, Xian, China, in 2009 and 2014 respectively. She is currently a professor with School of Information Engineering, Chang’an University, Xi'an, China. Her research interests include vision perception, autonomous vehicles, deep learning and intelligent transportation systems.
\end{IEEEbiography}

\begin{IEEEbiography}[{\includegraphics[width=1in,height=1.25in,clip,keepaspectratio]{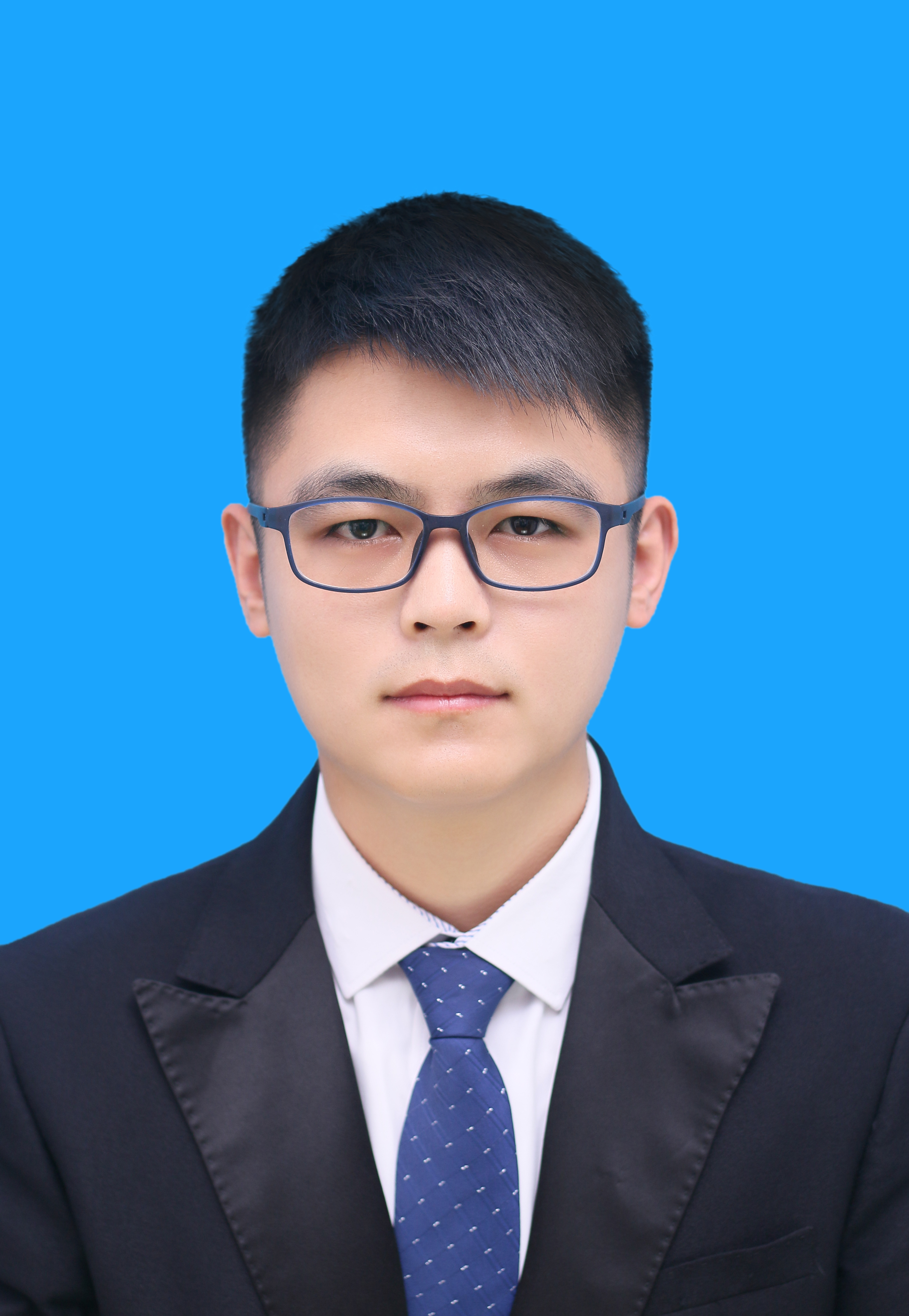}}]{Chao Li}
received the B.S. degree from Shandong University of Aeronautics in Binzhou, China, in 2020. He is currently working toward the Ph.D. degree in Traffic Information Engineering and Control at Chang’an University, Xi'an, China. His current research interests include vehicle trajectory prediction in complex dynamic traffic scenarios, and the generation of safety-critical trajectory scenarios.
\end{IEEEbiography}

\vspace{-20pt}

\begin{IEEEbiography}[{\includegraphics[width=1in,height=1.25in,clip,keepaspectratio]{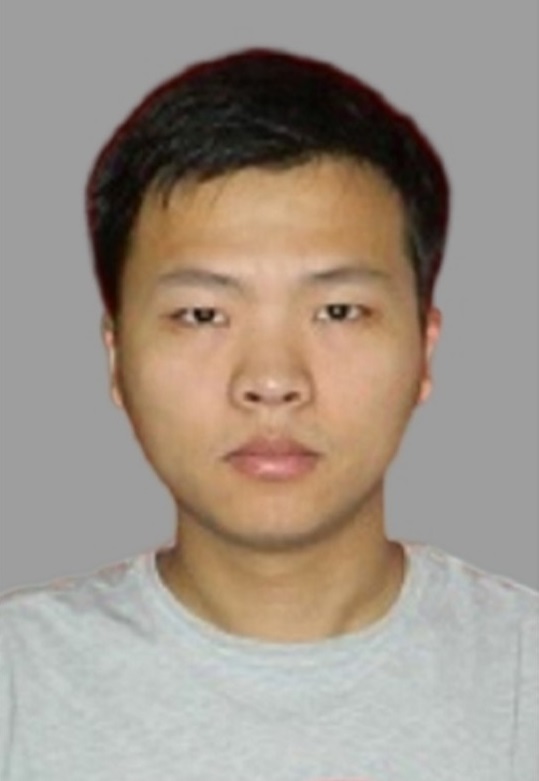}}]{Yang Wang}
(Member, IEEE) received the Ph.D. degree in control science and engineering from University of Science and Technology of China, in 2021. He is currently an associate professor with the School of Information Engineering, Chang’an University, Xi’an, China. His research interests include machine learning and image processing.
\end{IEEEbiography}

\vspace{-20pt}

\begin{IEEEbiography}[{\includegraphics[width=1in,height=1.25in,clip,keepaspectratio]{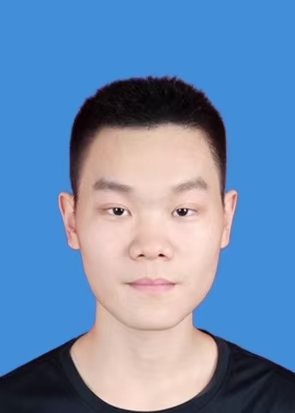}}]{Nan Yang}
(Member, IEEE) received the Ph.D. degree in control science and engineering from University of Science and Technology of China, in 2021. He is currently an associate professor with the School of Information Engineering, Chang’an University, Xi’an, China. His research interests include machine learning and image processing.
\end{IEEEbiography}

\vspace{-20pt}

\begin{IEEEbiography}[{\includegraphics[width=1in,height=1.25in,clip,keepaspectratio]{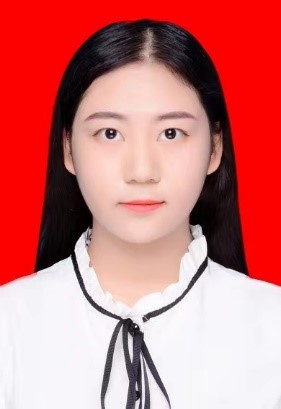}}]{Xing Fan}
received the B.S. degree in Measurement and Control Technology and Instruments from Chang’an University, Xi’an, China, in 2016, the M.S. degree in Computer Science and Technology from Chang’an University, Xi’an, China, in 2019. She is currently a lecturer with School of Electronic and Control Engineering, Chang’an University, Xi’an, China. Her research interests include vision perception, vehicle road collaboration and intelligent transportation systems.
\end{IEEEbiography}

\vspace{-20pt}

\begin{IEEEbiography}[{\includegraphics[width=1in,height=1.25in,clip,keepaspectratio]{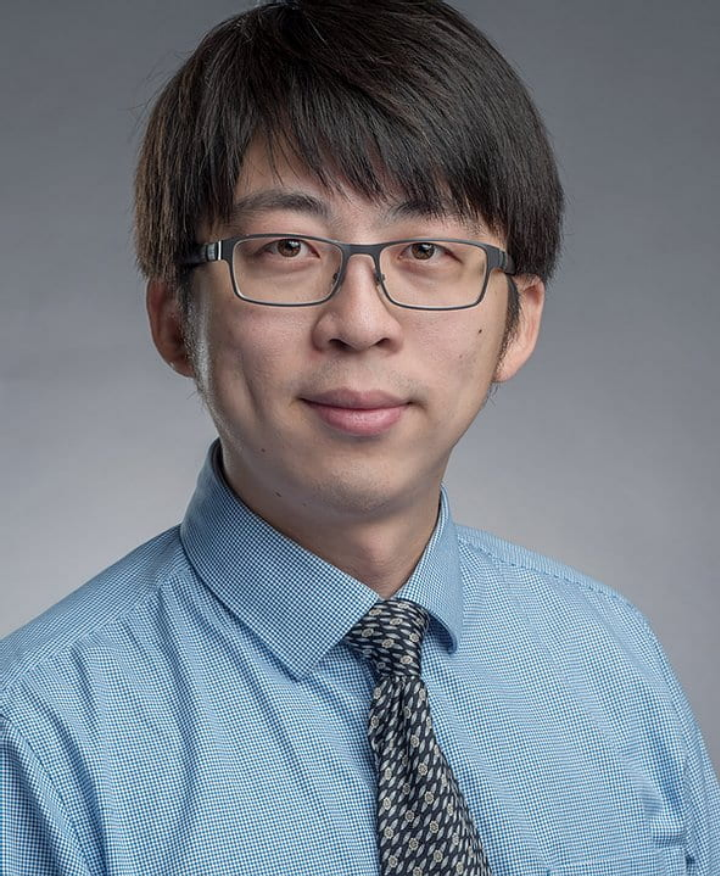}}]{Jiaqi Ma}
(Senior Member, IEEE) received the Ph.D. degree in transportation engineering from the University of Virginia, Charlottesville, VA, USA, in 2014. He is currently an Associate Professor with the Samueli School of Engineering, UCLA, where he is also the Faculty Lead of the New Mobility Program, Institute of Transportation Studies. His research interests include intelligent transportation systems, autonomous driving, and cooperative driving automation. He is a member of the TRB Standing Committee on Vehicle-Highway Automation, the TRB Standing Committee on Artificial Intelligence and Advanced Computing Applications, and the American Society of Civil Engineers (ASCE) Connected and Autonomous Vehicles Impacts Committee. He is the Co-Chair of the IEEE ITS Society Technical Committee on Smart Mobility and Transportation 5.0. He is the Editor-in-Chief of the IEEE OPEN JOURNAL OF INTELLIGENT TRANSPORTATION SYSTEMS.
\end{IEEEbiography}

\vspace{-20pt}

\begin{IEEEbiography}[{\includegraphics[width=1in,height=1.25in,clip,keepaspectratio]{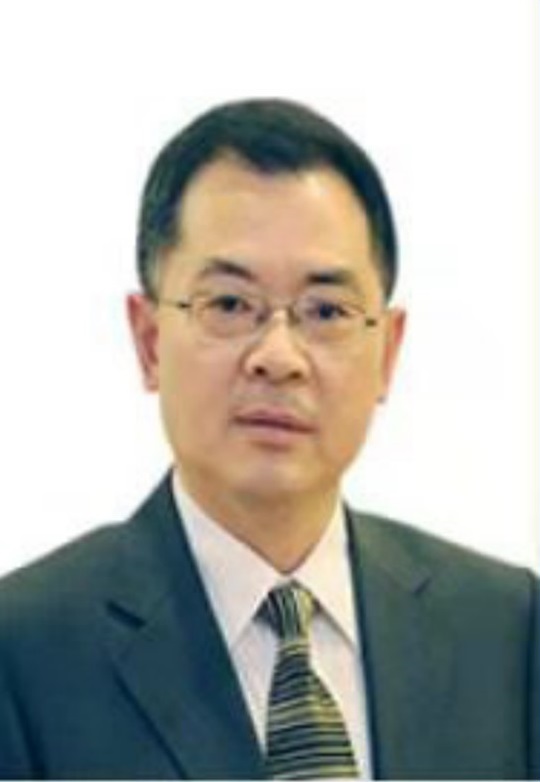}}]{Xiangmo Zhao}
 (Member, IEEE) received the B.S degree from Chongqing University, China, in 1987, and the M.S. and Ph.D. degrees from Chang’an University, China, in 2002 and 2005, respectively. He is currently a distinguished professor with School of Information Engineering, Chang’an University, Xi’an, China. His research interests include intelligent transportation systems, internet of vehicles, connected and autonomous vehicles testing technology.
\end{IEEEbiography}

\vfill

\end{document}